\documentclass[11pt]{article}

\usepackage[final]{acl}

\usepackage{times}
\usepackage{latexsym}
\usepackage[T1]{fontenc}
\usepackage[utf8]{inputenc}
\usepackage{microtype}
\usepackage{inconsolata}
\usepackage{graphicx}

\usepackage{amsmath}
\usepackage{amssymb}
\usepackage{booktabs}  
\usepackage{colortbl}  
\usepackage{xcolor}    
\usepackage{makecell}  
\usepackage{multirow}  
\usepackage{bm}
\usepackage{url}

\definecolor{lightgray}{gray}{0.9}
\usepackage{xspace} 
\newcommand{\ours}{\text{BEAVER}\xspace}

\title{\ours: A Training-Free Hierarchical Prompt Compression Method via Structure-Aware Page Selection}

\author{
Zhengpei Hu$^{1}$\thanks{\ Equal contribution.}, 
Kai Li$^{2,*}$, 
Dapeng Fu$^{3}$, 
Chang Zeng, 
Yue Li$^{1}$, 
Yuanhao Tang$^{1}$, 
Jianqiang Huang$^{1}$\thanks{\ Corresponding author.} \\
\\
$^{1}$School of Computer Technology and Application, Qinghai University \\
$^{2}$Tsinghua University \\
$^{3}$Ant Group Security and Intelligence Laboratory (SIL) \\
}

\begin{document}
\maketitle

\begin{abstract}
The exponential expansion of context windows in LLMs has unlocked capabilities for long-document understanding but introduced severe bottlenecks in inference latency and information utilization. Existing compression methods often suffer from high training costs or semantic fragmentation due to aggressive token pruning. In this paper, we propose \textit{\ours}, a novel training-free framework that shifts compression from linear token removal to structure-aware hierarchical selection. \ours maximizes hardware parallelism by mapping variable-length contexts into dense page-level tensors via dual-path pooling, and preserves discourse integrity through a hybrid planner combining semantic and lexical dual-branch selection with sentence smoothing. Extensive evaluations on four long-context benchmarks demonstrate that \ours achieves comparable performance to state-of-the-art (SOTA) methods like LongLLMLingua. Notably, on the RULER benchmark, \ours maintains high fidelity in multi-needle retrieval where baselines deteriorate. Regarding efficiency, \ours reduces latency by $26.4\times$ on 128k contexts, offering a scalable solution for high-throughput applications. Our code is available at \url{https://cslikai.cn/BEAVER/}.

\end{abstract}

\section{Introduction}

\begin{figure*}[t]
  \centering
  \includegraphics[width=0.88\linewidth]{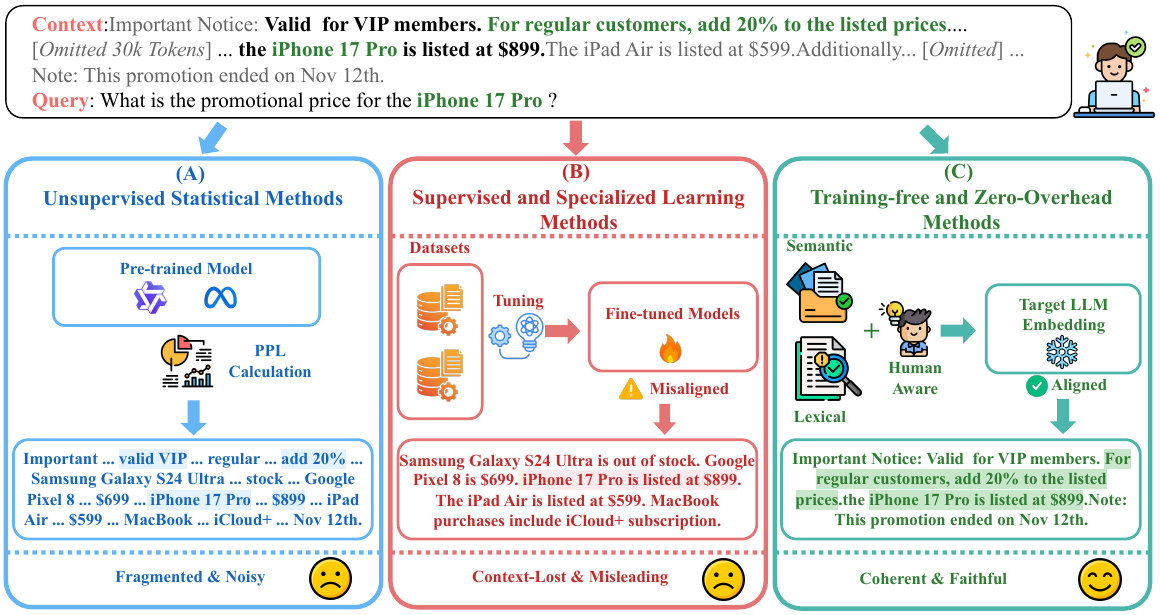}
  \caption{Comparison of different prompt compression paradigms for long-context LLMs.}
  \label{fig:example}
  \vspace{-15pt}
\end{figure*}

In recent years, the context window size of LLMs has expanded exponentially, evolving from the early 32k token configuration to the million-token scale supported by models such as Claude 4.5 \cite{anthropic2025claude45} and Gemini 3.0 \cite{DeepMind2025Gemini3Pro}. This advancement enables powerful capabilities for codebase analysis \cite{li2024loogle,li2025longcodeu} and global understanding across multiple long documents \cite{ma2024mmlongbench,deng2025longdocurl}, while simultaneously revealing several critical bottlenecks that hinder practical deployment. 

The first challenge is the "computation wall" in inference phase \cite{dao2022flashattention,zhao2025mofa}. While quantization \cite{lin2024awq} and system-level optimizations \cite{kwon2023efficient} reduce memory pressure, the $O(L^2)$ complexity of self-attention still causes prefill latency to surge with context length, leading to excessive time to first token and tail latency. The second challenge is the "diminishing returns" in information utilization \cite{sinha2025illusion}. Extending context windows does not yield proportional gains and often triggers the "Lost in the Middle" effect \cite{liu2024lost}, where models overlook key information. These challenges underscore that scaling length alone cannot ensure robustness, necessitating a new paradigm that balances efficiency and accuracy.

To address these challenges, prompt compression \cite{li2025prompt} offers an efficient solution by pruning redundancy while preserving semantics. Existing approaches fall into two paradigms (\ref{fig:example}): (i) Unsupervised statistical methods, such as Selective-Context \cite{li2023compressing} and the LongLLMLingua \cite{jiang2023llmlingua}, which filter tokens based on perplexity (PPL) or self-information from off-the-shelf models; and (ii) Supervised and specialized learning methods, like CPC \cite{liskavets2025prompt}, LLMLingua \cite{jiang2023llmlingua} and LLMLingua-2 \cite{pan2024llmlingua2}, which treat compression as a classification or ranking task using trained encoders to achieve higher precision.

However, two limitations hinder practical deployment. First, reliance on training incurs significant overhead and limits cross-model generalization, undermining the goal of lightweight inference. Second, unstructured token pruning disrupts semantic and syntactic coherence, resulting in fragmented contexts that impede long-sequence modeling.

\begin{figure*}[t]
  \centering
  \includegraphics[width=0.93\linewidth]{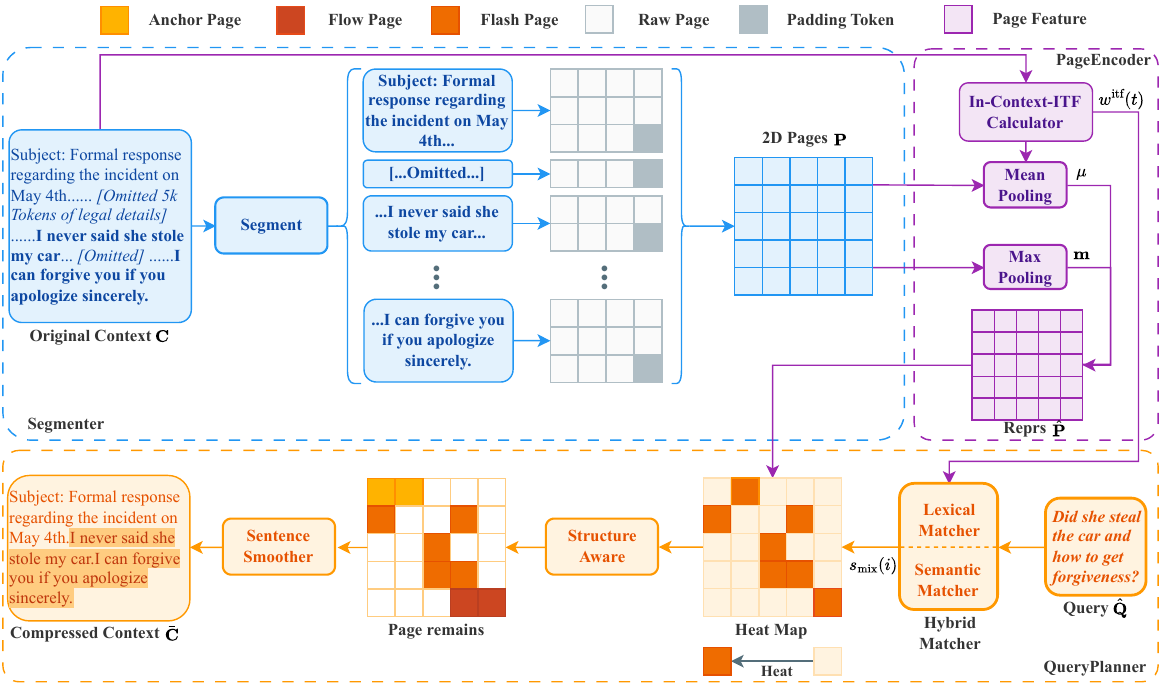}
  \caption{Overall pipeline of \textsc{\ours}. It comprises a Segmenter, a PageEncoder, and a QueryPlanner.}
  \label{fig:framework}
  \vspace{-15pt}
\end{figure*}

To address two these limitations, we propose \ours, a novel training-free framework that advances from linear token-wise pruning to structure-aware hierarchical selection, as shown in Figure~\ref{fig:framework}. The framework comprises three key components: a Segmenter that converts sequences into 2D page tensors to optimize GPU efficiency and preserve discourse; a PageEncoder that employs dual-path pooling to capture hierarchical features; and a QueryPlanner that integrates a triple structural prior (anchors, flow, flash) to mimic human cognition and suppress semantic drift \cite{aytes2025sketch}. Finally, a smoothing mechanism restores selections into coherent, high-fidelity compressed text.

Extensive evaluations on four long-context benchmarks (LongBench \cite{bai2024longbench}, ZeroSCROLLS \cite{shaham2023zeroscrolls}, RULER \cite{hsieh2024ruler}, and L-Eval \cite{an2024eval}) demonstrate that \ours achieves performance on par with or superior to SOTA methods like LongLLMLingua \cite{jiang2024longllmlingua} and LLMLingua-2 \cite{pan2024llmlingua2}. Notably, \ours dominates the challenging RULER benchmark, nearly doubling the performance of existing baselines. In terms of efficiency, it yields a $26.4\times$ speedup over LongLLMLingua at $128$k context. Moreover, as a training-free framework, \ours proves robust across diverse model scales (0.6B--32B), positioning it as a scalable solution for high-throughput long-document understanding.

\section{Related Work}

With the expansion of context windows in LLMs, prompt compression has become a vital strategy to mitigate inference costs by pruning redundant information while preserving core semantics \cite{li2025prompt}. Current research in hard prompt compression, which maintains discrete tokens for interpretability and compatibility, can be categorized into two primary paradigms.

The first paradigm consists of unsupervised statistical methods that rely on the perplexity or generation probabilities of off-the-shelf models without specialized training. For instance, Selective-Context \cite{li2023compressing} and the LongLLMLingua \cite{jiang2024longllmlingua} utilize self-information metrics to identify and prune redundant tokens or chunks. While efficient, these methods are often limited by their reliance on heuristic information-theoretic metrics. The second paradigm involves supervised and specialized learning methods, which optimize model weights specifically for compression objectives. LLMLingua and LLMLingua-2 \cite{jiang2023llmlingua, pan2024llmlingua2} reformulates compression as a token-level classification task via data distillation, while PCRL \cite{jung2024discrete} and TACO-RL \cite{shandilya2025taco} employ reinforcement learning for task-aware filtering. Additionally, works like CPC \cite{liskavets2025prompt}, AdaComp \cite{zhang2024adacomp}, and TCRA-LLM \cite{liu2023tcra} leverage specialized semantic encoders or embeddings to perform relevance filtering at the sentence or paragraph level, thereby mitigating local semantic fragmentation.

However, these methods face a dichotomy: learning-based approaches suffer from high deployment costs and limited transferability, while fine-grained token pruning disrupts syntactic coherence. Distinct from these methods, \ours introduces a training-free, structure-aware framework. By shifting from token-wise to segment-page selection, it effectively eliminates training dependencies while preserving discourse integrity, addressing the fragmentation issues prevalent in prior methods.

\section{Methods}
\label{sec:method}

\subsection{Overall Pipeline}
\label{subsec:overall}

Given a long-context sequence $\mathbf{C}\in \mathbb{Z}^{1\times L_c}$ and a query $\mathbf{Q}\in \mathbb{Z}^{1\times L_q}$, where $L_c$ and $L_q$ denote the number of tokens in the context and query respectively, our objective is to extract a compressed context $\tilde{\mathbf{C}}\in \mathbb{Z}^{1\times L_p}$ such that $L_p \ll L_c$, while preserving the key information required to derive the correct answer $\mathbf{A}$ as much as possible. As shown in Figure~\ref{fig:framework}, \textsc{\ours} consists of three components: Segmenter, PageEncoder, and QueryPlanner. 

Specifically, \ours applies compression only to the context preceding the query $\mathbf{Q}$. If $\mathbf{Q}$ is explicit, the prefix is compressed while $\mathbf{Q}$ is fully retained. If $\mathbf{Q}$ is implicit, it is defined as the final segment of the sequence, anchored to the nearest natural language boundary via a backward window to prevent truncation. 
The Segmenter then splits $\mathbf{C}$ into logical segments based on semantic delimiters (such as newlines or heading markers) and performs pagination to construct a structured two-dimensional page tensor $\mathbf{P}\in \mathbb{N}^{N\times M}$. On this basis, \ours introduces a dual-path pooling encoder (PageEncoder) equipped with a context-statistics-adaptive weighting mechanism, which jointly encodes the paginated token tensor $\mathbf{P}$ to obtain page-level representations $\hat{\mathbf{P}}\in \mathbb{R}^{N\times d}$, where $d$ denotes the feature dimension.  
The QueryPlanner executes the core compression by computing a hybrid semantic-lexical interaction score between $\hat{\mathbf{P}}$ and $\mathbf{Q}$. It constructs a saliency mask by integrating structural priors, such as anchor pages and flow pages, with high-scoring pages from the score distribution. Subsequently, sentence-level smoothing is applied to the salient pages. Finally, the discrete mask is mapped back to a continuous subsequence $\tilde{\mathbf{C}}$ for downstream LLM inference.

\subsection{Segmenter}
\label{subsec:segmenter}

The Segmenter is designed to efficiently map the variable-length sequence $\mathbf{C}$ into a 2D page matrix based on natural delimiters $\mathcal{D}$ (e.g., newlines). We first partition $\mathbf{C}$ into logical segments $\{\mathbf{c}_1,\dots,\mathbf{c}_K\}$. To construct the matrix, we employ a greedy pagination strategy with capacity $M$: multiple consecutive segments are packed into a single page if their cumulative length is within $M$, minimizing padding; longer segments are split across consecutive pages. The output is a page index tensor $\mathbf{P}$ (padded with $-1$), which enables efficient standard matrix multiplication while preserving local semantic boundaries.

\subsection{PageEncoder}
\label{subsec:pageencoder}

To enable QueryPlanner to efficiently perform coarse-grained selection, \ours uses a dual-path pooling encoder named PageEncoder. It captures both global semantics and salient local features through weighted average and max pooling while utilizing a context-adaptive weighting mechanism to mitigate semantic collapse caused by redundant tokens.

Specifically, given the LLM embedding $E:\mathbb{N}\rightarrow\mathbb{R}^{d}$, we first map the token sequence to token features $\mathbf{H}\in\mathbb{R}^{L_c\times d}$, where the feature of the $\ell$-th token is $\mathbf{h}_\ell=E(C_\ell)$. Using the page index tensor $\mathbf{P}$ generated by the Segmenter, we rearrange the token features into a page-level tensor $\mathbf{X}\in\mathbb{R}^{N\times M\times d}$, where the feature at position $(i,j)$ is defined as
\begin{equation}
    \mathbf{x}_{i,j}=
    \begin{cases}
    \mathbf{h}_{\ell}, & \text{if } P_{i,j}=\ell\ge0,\\[2pt]
    \mathbf{0}, & \text{if } P_{i,j}=-1.
    \end{cases}
\end{equation}

To down-weight frequent but uninformative tokens, we compute an in-context Inverse Term Frequency (ITF) score. Let $\mathrm{tf}(t)$ be the token frequency within the context, the weight is defined as:
\begin{equation}
    w^{\text{itf}}(t)=\text{Norm}\left(\log\Bigl(1+\frac{L_c+L_q}{1+\mathrm{tf}(t)}\Bigr)\right),
\end{equation}
where $\text{Norm}(\cdot)$ denotes min-max normalization to $[0,1]$. The effective attention weight for position $(i,j)$ is then $\omega_{i,j}= B_{i,j}\cdot w^{\text{itf}}(C_{P_{i,j}})$, with binary mask $B_{i,j}=\mathbb{I}[P_{i,j}\ge0]$.

We then extract complementary representations via two pooling paths. The weighted average pooling aggregates global semantics:
\begin{equation}
    \boldsymbol{\mu}_i=\frac{\sum_{j=1}^{M}\omega_{i,j}\mathbf{x}_{i,j}}{\sum_{j=1}^{M}\omega_{i,j}+\varepsilon},
\end{equation}
where $\varepsilon$ ensures numerical stability. Simultaneously, the max pooling path captures salient local activations (e.g., rare keywords). We mask invalid positions with a large negative constant $\beta$ to ensure numerical correctness:
\begin{equation}
\mathbf{m}_i = \max_{1\le j\le M}\bigl(B_{i,j}\mathbf{x}_{i,j}+(1-B_{i,j})\cdot\beta\bigr).
\end{equation}
The final representation is a linear fusion
\begin{equation}
\hat{\mathbf{p}}_i=\gamma\boldsymbol{\mu}_i+(1-\gamma)\mathbf{m}_i,
\end{equation}
where $\gamma$ is fusion weights. All page representations are stacked into a matrix $\hat{\mathbf{P}}\in\mathbb{R}^{N\times d}$.

Similarly, for short queries (length $L_q < 4$), we adopt a dense retrieval approach \cite{karpukhin2020dense} to obtain a unified semantic vector $\mathbf{q}\in\mathbb{R}^{d}$. For longer queries ($L_q \ge 4$), we employ a late interaction strategy \cite{khattab2020colbert}, retaining fine-grained token representations to handle complex matching requirements.

\subsection{QueryPlanner}
\label{subsec:queryplanner}

The QueryPlanner identifies salient pages from the encoded set $\hat{\mathbf{P}}$ to construct a compressed context. To capture both semantic relevance and exact answer spans, it integrates dense retrieval and lexical overlap within a unified space, incorporated with structural priors.

Given the query representation $\mathbf{Q}=\{\mathbf{q}_k\}_{k=1}^K$ (where $K=1$ for short queries) and the set of page vectors $\hat{\mathbf{P}}=\{\hat{\mathbf{p}}_i\}_{i=1}^N$, we compute a semantic score via a weighted cosine similarity. To handle multi-vector queries effectively, we define:
\begin{equation}
s_{\mathrm{sem}}(i)=\sum_{k=1}^{K}w^{\text{itf}}_{q_k} \frac{\hat{\mathbf{p}}_i\cdot\mathbf{q}_k}{\|\hat{\mathbf{p}}_i\|_2\cdot\|\mathbf{q}_k\|_2},
\end{equation}
where $w^{\text{itf}}_{q_k}$ is the ITF score of $\mathbf{q}_k$. This mechanism naturally assigns higher importance to discriminative terms (e.g., entities) over common stop words.

Complementarily, the lexical branch highlights exact token overlaps. Let $\mathcal{T}_{\mathcal{Q}}$ be the query token set and $\mathcal{P}_i$ be the tokens in page $i$. We aggregate the importance of overlapping tokens:
\begin{equation}
s_{\mathrm{lex}}(i)
=
\sum_{\ell\in\mathcal{P}_i}
\mathbb{I}[C_\ell\in\mathcal{T}_{\mathcal{Q}}]\cdot w^{\text{itf}}_{C_\ell}.
\end{equation}
Both scores are min-max normalized to $[0,1]$ over all $N$ pages. The final relevance score is a linear fusion $s_{\mathrm{mix}}(i) = \lambda s_{\mathrm{sem}}(i) + (1-\lambda) s_{\mathrm{lex}}(i)$, where $\lambda$ is one hyperparameter that control the relative importance of semantic and lexical alignment. Tasks emphasizing semantic reasoning may assign a larger $\lambda$, while those containing many code snippets or identifiers may benefit from a smaller $\lambda$.  

After computing the scores $s_{\mathrm{mix}}(i)$, QueryPlanner incorporates structural priors to enhance stability and discourse coherence. We first identify the query anchor $p_{\mathrm{qry}}$ (the page containing the query start) to enforce causal constraints ($i \le p_{\mathrm{qry}}$). The selection set is composed of three distinct subsets: (1) Anchors: The initial $k_{\mathrm{anc}}$ pages are always preserved to retain global metadata (e.g., title, introductory definitions). (2) Flow: To mimic human working memory~\cite{johnson2010mental}, we select a contiguous window size $w_{\mathrm{flow}}$ immediately preceding the query, defined as indices $i \in [\max(1, p_{\mathrm{qry}}-w_{\mathrm{flow}}), p_{\mathrm{qry}}]$. (3) Flash: From the remaining candidates (excluding Anchors and Flow), we select the top-$k_{\mathrm{flash}}$ pages with the highest $s_{\mathrm{mix}}(i)$ to capture distant but critical evidence.

\begin{figure}[t]
  \centering
  \includegraphics[width=0.8\linewidth]{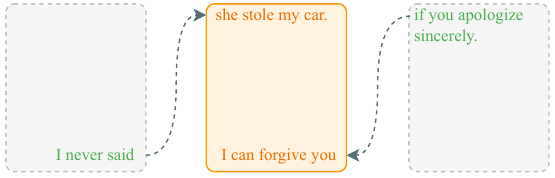}
  \caption{Illustration of sentence-level smoothing. To ensure syntactic completeness, the initially retrieved span (orange solid box) is extended outward to the nearest sentence boundaries (dashed boxes).}
  \label{fig:smooth}
  \vspace{-15pt}
\end{figure}

Finally, the selected page indices are mapped back to token spans $[a_r, b_r]$. As illustrated in Figure~\ref{fig:smooth}, we apply sentence-level smoothing: start $a_r$ and end $b_r$ indices are extended outward to the nearest sentence boundaries to preserve syntactic completeness. Overlapping spans are merged, and the result is concatenated to form the final input $\tilde{\mathbf{C}}$. 

\begin{table*}[t]
    \centering
    \setlength{\tabcolsep}{3.5pt}

    \resizebox{\textwidth}{!}{
        \begin{tabular}{l | cccccc >{\columncolor{lightgray}}c | cc | cc | >{\columncolor{lightgray}}c | cc | cc}
            \toprule
            \multirow{2}{*}{\textbf{Methods}} & \multicolumn{11}{c|}{\textbf{LongBench}} & \multicolumn{5}{c}{\textbf{ZeroSCROLLS}} \\
            \cmidrule(lr){2-12} \cmidrule(lr){13-17}
            & SingleDoc & MultiDoc & Summ. & FewShot & Synth. & Code & \textbf{AVG} & Tokens & $1/\tau$ & Lat. & Spd. & \textbf{AVG} & Tokens & $1/\tau$ & Lat. & Spd. \\
            \midrule
            \hline
            \multicolumn{17}{c}{\textit{2,000-token constraint}} \\

            \midrule
            \multicolumn{17}{l}{\textit{(1) Unsupervised Statistical Methods}} \\
            Selective-Context & 16.2 & 34.8 & 24.4 & 15.7 & 8.4 & 49.2 & 24.8 & 1,925 & 5x & 47.1 & 0.3x & 19.4 & 1,865 & 5x & 47.5 & 0.3x \\
            LongLLMLingua & \underline{39.0} & \textbf{42.2} & \textbf{27.4} & \textbf{69.3} & \textbf{53.8} & 56.6 & \textbf{48.0} & 1,809 & 6x & 6.1 & 2.6x & 32.7 & 1,753 & 6x & 5.2 & 2.3x \\

            \midrule
            \multicolumn{17}{l}{\textit{(2) Supervised and Specialized Learning Methods}} \\
            SBERT & 33.8 & 35.9 & \underline{25.9} & 23.5 & 18.0 & 17.8 & 25.8 & 1,947 & 5x & 4.8 & 3.4x & 20.5 & 1,773 & 6x & 4.1 & 3.0x \\
            OpenAI & 34.3 & \underline{36.3} & 24.7 & 32.4 & \underline{26.3} & 24.8 & 29.8 & 1,991 & 5x & 10.4 & 1.5x & 20.6 & 1,784 & 5x & 9.9 & 1.2x \\
            LLMLingua & 22.4 & 32.1 & 24.5 & 61.2 & 10.4 & \underline{56.8} & 34.6 & 1,950 & 5x & 5.9 & 2.6x & 27.2 & 1,862 & 5x & 4.8 & 2.5x \\
            LLMLingua-2-small & 29.5 & 32.0 & 24.5 & 64.8 & 22.3 & 56.2 & 38.2 & 1,891 & 5x & \underline{3.3} & \underline{4.7x} & \underline{33.3} & 1,862 & 5x & \underline{2.6} & \underline{4.7x} \\
            LLMLingua-2 & 29.8 & 33.1 & 25.3 & \underline{66.4} & 21.3 & \textbf{58.9} & 39.1 & 1,954 & 5x & 3.7 & 4.2x & \textbf{33.4} & 1,898 & 5x & 3.0 & 4.1x \\

            \midrule
            
            \multicolumn{17}{l}{\textit{(3) Training-free and Zero-Overhead Methods}} \\
            \rowcolor{blue!5} 
            \textbf{\ours (ours)} & \textbf{40.7} & 37.6 & 22.1 & 57.4 & 38.2 & 57.1 & \underline{42.2} & 1,982 & 5x & \textbf{3.0} & \textbf{5.2x} & 32.0 & 1,878 & 5x & \textbf{2.5} & \textbf{4.9x} \\

            \midrule
            \hline
            \multicolumn{17}{c}{\textit{3,000-token constraint}} \\
            \hline

            \multicolumn{17}{l}{\textit{(1) Unsupervised Statistical Methods}} \\
            Selective-Context & 23.3 & \underline{39.2} & 25.0 & 23.8 & 27.5 & 53.1 & 32.0 & 3,328 & 3x & 50.6 & 0.3x & 20.7 & 3,460 & 3x & 54.2 & 0.2x \\
            LongLLMLingua & \underline{40.7} & \textbf{46.2} & \textbf{27.2} & \textbf{70.6} & \textbf{53.0} & 55.2 & \textbf{48.8} & 3,283 & 3x & 10.0 & 1.6x & 33.0 & 3,412 & 3x & 8.2 & 1.5x \\
            
            \midrule
            \multicolumn{17}{l}{\textit{(3) Supervised and Specialized Learning Methods}} \\
            SBERT & 35.3 & 37.4 & 26.7 & 63.4 & \underline{51.0} & 34.5 & 41.4 & 3,399 & 3x & 7.7 & 2.0x & 24.0 & 3,340 & 3x & 5.9 & 2.1x \\
            OpenAI & 34.5 & 38.6 & \underline{26.8} & 63.4 & 49.6 & 37.6 & 41.7 & 3,421 & 3x & 13.3 & 1.2x & 22.4 & 3,362 & 3x & 11.7 & 1.0x \\
            LLMLingua & 31.8 & 37.5 & 26.2 & 67.2 & 8.3 & 53.2 & 37.4 & 3,421 & 3x & 9.2 & 1.7x & 30.7 & 3,366 & 3x & 7.4 & 1.7x \\
            LLMLingua-2-small & 35.5 & 38.1 & 26.2 & 67.5 & 23.9 & \underline{60.0} & 41.9 & 3,278 & 3x & \underline{3.9} & \underline{4.0x} & \underline{33.4} & 3,089 & 3x & \underline{3.0} & \underline{4.1x} \\
            LLMLingua-2 & 35.5 & 38.7 & 26.3 & \underline{69.6} & 21.4 & \textbf{62.8} & 42.4 & 3,392 & 3x & 4.3 & 3.6x & \textbf{33.5} & 3,206 & 3x & 3.5 & 3.5x \\

            \midrule
            \multicolumn{17}{l}{\textit{(3) Training-free and Zero-Overhead Methods}} \\
            \rowcolor{blue!5}
            \textbf{\ours (ours)} & \textbf{42.3} & 39.0 & 22.6 & 60.8 & 43.7 & 56.7 & \underline{44.2} & 3,289 & 3x & \textbf{3.5} & \textbf{4.5x} & 32.4 & 3,319 & 3x & \textbf{2.9} & \textbf{4.2x} \\

            \midrule
            \hline
            \multicolumn{17}{l}{\textit{Reference Baselines}} \\
            Original Prompt & 39.7 & 38.7 & 26.5 & 67.0 & 37.8 & 54.2 & 44.0 & 10,295 & - & 15.6 & - & 32.5 & 9,788 & - & 12.2 & - \\
            Zero-Shot & 15.6 & 31.3 & 15.6 & 40.7 & 1.6 & 36.2 & 23.5 & 214 & 48x & 1.6 & 9.8x & 12.8 & 32 & 306x & 1.0 & 12.2x \\
            \bottomrule
        \end{tabular}
    }
    \caption{Performance comparison on LongBench and ZeroSCROLLS benchmarks with a 2,000 and 3,000-token budget. Best results are bolded, and second best are underlined. 
    "Tokens" reports the average number of tokens remaining after compression, and "$1/\tau$" is the average compression ratio. "Lat." and "Spd." denote Latency and Speedup, respectively.}
    \label{tab:main_comparison_longbench_zero}
    \vspace{-15pt}
\end{table*}

\begin{table*}[t]
    \centering
    \renewcommand{\arraystretch}{1.15}
    \setlength{\tabcolsep}{3.5pt} 

    \resizebox{0.9\textwidth}{!}{
        \begin{tabular}{l | ccc | cccccc | c | cc | c | c}
            \toprule
            \multirow{2}{*}{\textbf{Method}} & 
            \multicolumn{3}{c|}{\textbf{Single}} & 
            \multicolumn{6}{c|}{\textbf{Multi}} &  
            \textbf{FWE} & 
            \multicolumn{2}{c|}{\textbf{QA}} & 
            \multirow{2}{*}{\textbf{AVG}} &
            \multirow{2}{*}{\textbf{\shortstack{Avg\\Tokens}}} \\
            
            \cmidrule(lr){2-4} \cmidrule(lr){5-10} \cmidrule(lr){11-11} \cmidrule(lr){12-13}
            
            & S-1 & S-2 & S-3 & Key-1 & Key-2 & Key-3 & Val & Qry & VT & Freq & QA-1 & QA-2 & & \\
            \midrule
            
            LLMLingua  & \textbf{100.0} & 5.0 & 4.0 & 7.0 & 6.0 & \underline{12.0} & 6.0 & 4.8 & 59.4 & 90.7 & 15.0 & 37.0 & 28.9 & 3,230 \\
            LongLLMLingua  & \textbf{100.0} & 6.0 & 4.0 & 9.0 & 2.0 & 3.0 & 7.8 & 7.3 & \underline{60.4} & \underline{91.3} & 19.0 & 36.0 & 28.8 & 3,179 \\
            LLMLingua-2  & \underline{27.0} & \underline{86.0} & \underline{27.0} & \underline{72.0} & \underline{11.0} & 2.0 & \underline{80.3} & \underline{82.3} & 3.4 & \textbf{95.3} & \underline{45.0} & \underline{43.0} & \underline{47.9} & 3,132 \\
            
            \rowcolor{blue!5} 
            \textbf{\ours (ours)} & \textbf{100.0} & \textbf{100.0} & \textbf{100.0} & \textbf{99.0} & \textbf{88.0} & \textbf{41.0} & \textbf{99.8} & \textbf{99.8} & \textbf{78.2} & 93.7 & \textbf{50.0} & \textbf{55.0} & \textbf{83.7} & 3,198 \\
            
            \bottomrule
        \end{tabular}
    }
    \caption{Performance of \ours and baselines on RULER (16k context, 3k budget). "S": Single-needle; "Multi": Multi-needle (Keys/Values) \& Variable Tracking (VT); "FWE": Frequent Word Extraction; "QA": Question Answering.}
    \label{tab:main_results_ruler}
\end{table*}

\begin{table*}[t]
    \centering
    \setlength{\tabcolsep}{4pt}
    \resizebox{0.9\textwidth}{!}{
        \begin{tabular}{l | cccccc >{\columncolor{lightgray}}c | cc}
            \toprule
            \textbf{Methods} 
             & Coursera QA & QuALITY & SFictionQA & TPO & LongFQA & Legal Contract QA & \textbf{AVG} 
             & \textbf{Tokens} & \bm{$1/\tau$}\\
            \midrule
            LLMLingua 
            & 58.9 & 50.0 & 60.9 & 65.1 & 34.3 & 21.9 & 48.5 
            & 2,122 & 4$\times$ \\
            
            LongLLMLingua
            & 62.2 & 51.0 & 65.6 & 71.0 & 37.7 & 21.3 & 51.5 
            & 2,196 & 4$\times$ \\
            
            LLMLingua-2 
            & \underline{64.4} & \underline{55.0} & 66.4 & \underline{73.2} & \textbf{45.0} & 23.7 & 54.6 
            & 2,110 & 4$\times$ \\

            Simply-BEAVER
            & 63.1 & 52.9 & \underline{74.2} & 72.5 & 39.7 & \underline{27.2} & \underline{54.9} 
            & 2,123 & 4$\times$ \\
            
            \midrule
            \rowcolor{blue!5} 
            \textbf{\ours (ours)} 
            & \textbf{64.4} & \textbf{56.9} & \textbf{76.6} & \textbf{74.0} & \underline{44.7} & \textbf{28.8} & \textbf{57.6} 
            & 2,180 & 4$\times$ \\
            
            \bottomrule
        \end{tabular}
    }
    \caption{Out-of-domain evaluation on general long-context scenarios with a 2,000-token budget.}
    \label{tab:main_comparison_leval}
    \vspace{-15pt}
\end{table*}

\section{Experiments}

\subsection{Experimental Setup}

\paragraph{Benchmarks.} To comprehensively evaluate the effectiveness and robustness of \ours, we conducted experiments on four diverse and challenging long-context benchmarks. We first adopted LongBench~\cite{bai2024longbench} and ZeroSCROLLS~\cite{shaham2023zeroscrolls}, which cover multiple tasks including single-document and multi-document question answering, summarization, and few-shot learning, providing a holistic view of general long-context capabilities. We then used the high-quality synthetic benchmark RULER~\cite{hsieh2024ruler} to test long-context understanding under varying context lengths from 16k to 128k, including tasks such as multi-needle retrieval and variable tracking, which precisely assess the model’s ability to retain fine-grained information. Finally, we further evaluated the generalization performance of \ours on out-of-domain tasks using L-Eval~\cite{an2024eval}, which enforces strict token-length constraints. More detailed dataset descriptions and evaluation metrics are provided in Appendix~\ref{appendix:datasets}.  


\paragraph{Baseline Models.} To ensure a fair comparison, we evaluated \ours aagainst several prompt compression methods. The baselines are grouped into two categories: (1) Unsupervised statistical methods such as Selective-Context~\cite{li2023compressing} and LongLLMLingua~\cite{jiang2024longllmlingua}, which rely on intrinsic metrics such as perplexity or self-information from off-the-shelf models to identify redundancy without specialized training. (2) Supervised and specialized learning methods, which utilize models explicitly optimized for compression tasks or leverage specialized semantic encoders such as LLMLingua~\cite{jiang2023llmlingua} and LLMLingua-2~\cite{pan2024llmlingua2}. Additionally, we included embedding-based retrieval approaches, like SBERT~\cite{reimers2019sentence} and OpenAI Embeddings~\cite{openai_embeddings}, within this paradigm as they employ trained encoders to perform semantic relevance filtering. Baseline configurations and implementation details are available in Appendix~\ref{appendix:baselines}. 

\paragraph{Implementation Details.}
For the inference backend, we followed existing methods \cite{jiang2024longllmlingua,pan2024llmlingua2} and uniformly adopted \texttt{gpt-3.5-turbo-instruct}\footnote{\url{https://platform.openai.com/docs/models/gpt-3.5-turbo-instruct}} as the large language model for all downstream task evaluations. The PageEncoder utilized embeddings from \texttt{Qwen3-8B}\footnote{\url{https://huggingface.co/Qwen/Qwen3-8B}}.
Regarding hyperparameters, we set the page size $M=64$, fusion weight $\gamma=0.7$, and scoring parameter $\lambda=0.7$.
Structural priors were fixed at $k_{\mathrm{anc}}=w_{\mathrm{flow}}=4$, with the flash set size $k_{\mathrm{flash}}$ dynamically adapted to fully utilize the remaining token budget.
Our evaluations strictly adhered to benchmark-specific context budgets and employ official standard prompt templates to ensure fair comparison.
All latency and throughput metrics were measured on NVIDIA A100 (80GB) GPUs.

\subsection{Comparison with SOTA Methods}
\label{subsec:comparison}

\subsubsection{Performance Analysis}
To demonstrate the advantages of \ours, we compared it with SOTA methods on four different benchmarks: LongBench~\cite{bai2024longbench}, ZeroSCROLLS~\cite{shaham2023zeroscrolls}, RULER \cite{hsieh2024ruler}, and L-Eval \cite{an2024eval}, under strict token budgets of 2,000 and 3,000 tokens.

As shown in Table~\ref{tab:main_comparison_longbench_zero}, \ours consistently outperforms the latest learning-based LLMLingua-2~\cite{pan2024llmlingua2} on LongBench, establishing a new SOTA of 40.7 on single-document QA. While LongLLMLingua \cite{jiang2024longllmlingua} excels on synthetic tasks, it suffers from high computational costs due to its coarse-to-fine re-ranking mechanism. In contrast, \ours is fully training-free and achieves comparable results on ZeroSCROLLS (average 32.0 vs. 32.7, see Appendix~\ref{appendix:details_zero} for detailed analysis) with significantly reduced latency. Specifically, \ours achieves the lowest latency (3.0s) and highest speedup (5.2$\times$), validating that our segment-page hierarchical design effectively minimizes inference overhead.

To evaluate the capability of locating specific information within large contexts, we conducted experiments on the RULER benchmark (Table~\ref{tab:main_results_ruler}). Existing compression methods suffer from severe performance degradation in this setting. For example, LLMLingua and LongLLMLingua fail to retrieve key information in multi-needle tasks and achieve only single-digit scores on S-2 and S-3. In contrast, \ours exhibits strong robustness, maintaining 100.0\% accuracy on all single-needle tasks and achieving an average score of 83.7, which substantially surpasses the second-best method LLMLingua-2 (47.9). This capability is further corroborated by the Needle-in-a-Haystack visualizations in Appendix~\ref{appendix:needle}. As shown in Figures~\ref{fig:single-needle} and \ref{fig:multi-needle}, \ours attains near-perfect recall on Single-Needle (100\%) and Multi-Needle (99\%) tasks, whereas LongLLMLingua fails almost completely ($<10\%$) and LLMLingua-2 shows instability (86\% and 72\% respectively). This sharp contrast highlights the advantage of our PageEncoder and QueryPlanner, which alleviate the "lost in the middle" issue~\cite{liu2024lost} by preserving local semantic structure and leveraging in-context ITF to capture rare but crucial tokens that are often discarded by probability-based pruners.

To further assess the generalization ability of \ours under unseen distributions, we report results on the L-Eval benchmark in Table~\ref{tab:main_comparison_leval}. We also extend this evaluation to the open-weights Qwen3-8B model in Appendix~\ref{appendix:leval_results} to demonstrate robust generalization across model families. Despite being training-free, \ours demonstrates superior adaptability compared to learning-based baselines trained on extensive datasets. \ours achieved the highest average score of 57.6, outperforming LongLLMLingua (51.5) and LLMLingua-2 (54.6). \ours ranks first on 4 out of 6 datasets, with notable gains on SFictionQA and Legal Contract QA. These results confirm that our structure-aware sentence smoothing mechanism effectively preserves discourse coherence, which is crucial for complex reasoning tasks in domains such as law and literature. We provide concrete examples comparing the compression outputs of LongLLMLingua, LLMLingua-2, and \ours in Appendix~\ref{appendix:qualitative_comparison} to illustrate how token-level fragmented compression can disrupt narrative consistency.

\subsubsection{Efficiency Analysis}

Efficiency is critical for real-world deployment, particularly as context lengths exceed 100k tokens. To evaluate the computational scalability, we conducted a latency analysis and compared with the LongLLMLingua and LLMLingua-2 families (including standard and small variants). Experiments were performed under different context lengths ranging from 16k to 128k, with the target output length fixed at 3,000 tokens.

\begin{figure}[t]
    \centering
    \includegraphics[width=\linewidth]{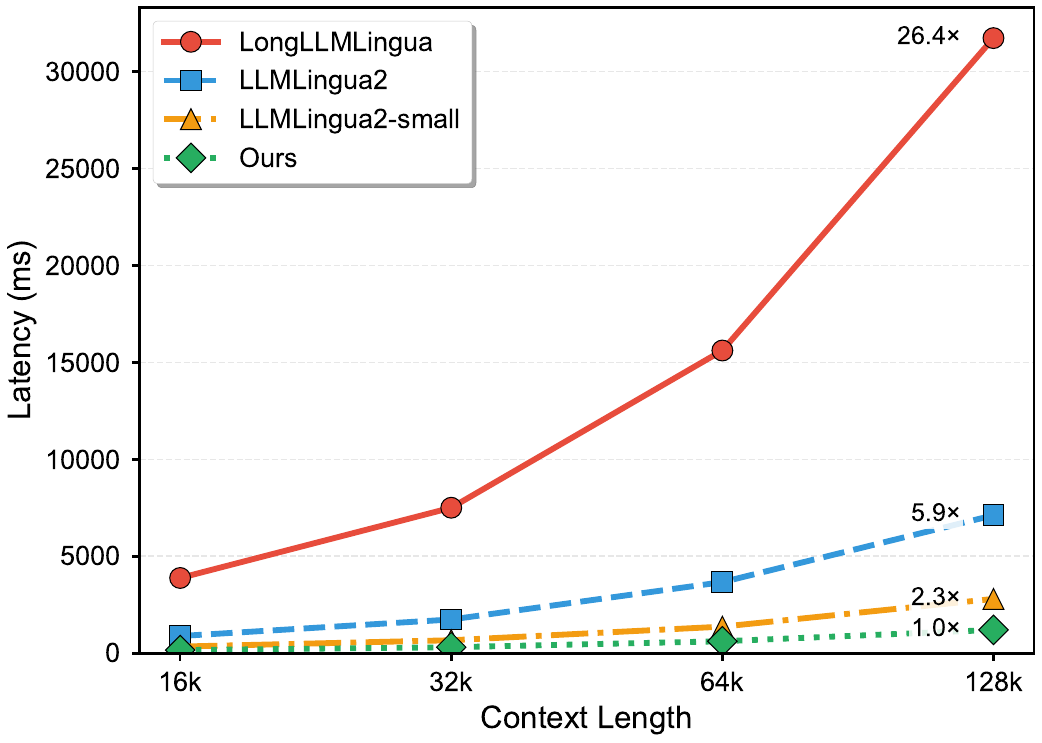}
    \caption{Comparison of inference latency versus context length for different compression methods.}
    \label{fig:latency_comparison}
    \vspace{-15pt}
\end{figure}

Quantitative results are shown in Fig.~\ref{fig:latency_comparison}. \ours exhibits a clear efficiency advantage across all context lengths. For a 128k-token ultra-long context, our method completes the compression in only 1.20 s, while the coarse-to-fine LongLLMLingua requires about 31.7 s, corresponding to a 26.4$\times$ speedup. Even compared with the highly optimized distilled models LLMLingua-2 and its lightweight variant LLMLingua-2-small, \ours still achieves 5.9$\times$ and 2.3$\times$ speedup, respectively. Moreover, the latency of \ours grows linearly with a significantly flatter slope than all baselines. This suggests that our segment-page hierarchical design effectively maximizes parallelism, avoiding the computational bottlenecks typical of sequential processing.

\subsection{Ablation Study}
\label{subsec:ablation_comp}

To evaluate the effectiveness of each component, we conducted detailed ablations on the QA subset of LongBench. All experiments used a 2,000-token compression budget. Results are summarized in Table~\ref{tab:component_ablation}, covering segmentation granularity, PageEncoder representations, QueryPlanner scoring, and hierarchical page selection strategies.

\textbf{Impact of Segmentation Granularity.} We first examined the physical page size $M$. As shown in Table~\ref{tab:component_ablation}, using overly small pages ($M=32$) reduces the average score by 1.8 points, since fine granularity fragments contiguous local contexts. Conversely, using overly large pages ($M=128$) decreases performance by 2.4 points, indicating that coarse pages introduce substantial background noise, diluting the density of useful information.

\textbf{Effectiveness of Encoder and Planner Designs.} Regarding PageEncoder, removing max or mean pooling yields a $\sim$2.6 point drop, confirming their complementarity. Ablating the multi-vector query strategy causes a significant degradation ($-2.9$), suggesting that fine-grained query interactions are critical for capturing complex constraints. We also verify the robustness of the backbone embedding model: replacing the default Qwen3-8B Embedding with Llama3-8B Embedding yields a negligible fluctuation ($\Delta=-0.3$), indicating that our framework generalizes well across different model families. For QueryPlanner, removing lexical matching leads to the most severe drop ($\Delta=-6.0$), highlighting the necessity of exact surface overlap. Removing semantic matching also incurs a 3.1 point loss. Furthermore, removing the in-context ITF score reduces performance by 2.7 points, validating that unsupervised context-aware weighting effectively suppresses high-frequency noise. Sentence smoothing provides a modest gain ($+1.6$), proving vital for repairing truncated sentence boundaries in multi-document tasks.

\textbf{Significance of Hierarchical Selection.} Finally, we analyzed the page selection strategy. Using only \textit{Flash} achieves a score of 36.4 but remains below the full model. Relying solely on \textit{Flow} causes a sharp drop to 31.1, indicating that local context alone is insufficient. Keeping only \textit{Anchor} leads to a collapse in reasoning performance (17.5). These results validate the hybrid design of \ours: Flash retrieves highly relevant evidence, while Anchor and Flow ensure global instruction following and local coherence.

\begin{table}[t]
\centering
\small
\resizebox{\linewidth}{!}{
\begin{tabular}{lcccc}
\toprule
\textbf{Configuration} & \textbf{S-Doc} & \textbf{M-Doc} & \textbf{Avg.} & \textbf{$\Delta$} \\
\midrule
\rowcolor{green!10}\textbf{\ours} & \textbf{40.7} & \textbf{37.6} & \textbf{39.2} & -- \\
\midrule
\rowcolor{gray!15}\multicolumn{5}{l}{\textbf{Segmenter}} \\
Page Size $M$=32 & 38.5 & 36.3 & 37.4 & {\color{red}-1.8} \\
Page Size $M$=128 & 37.5 & 36.1 & 36.8 & {\color{red}-2.4} \\
\rowcolor{gray!15}\multicolumn{5}{l}{\textbf{PageEncoder}} \\
\quad w/o Max-Pooling & 39.2 & 33.7 & 36.5 & {\color{red}-2.7} \\
\quad w/o Mean-Pooling & 39.2 & 34.0 & 36.6 & {\color{red}-2.6} \\
\quad w/o Multi-Token Query & 38.3 & 34.3 & 36.3 & {\color{red}-2.9} \\
\quad LLaMA3-8B Embedding & 39.4 & 38.2 & 38.8 & {\color{orange}-0.3} \\
\rowcolor{gray!15}\multicolumn{5}{l}{\textbf{QueryPlanner}} \\
\quad w/o ITF Score & 35.7 & 37.3 & 36.5 & {\color{red}-2.7} \\
\quad Semantic Only & 36.5 & 29.9 & 33.2 & {\color{red!80!black}\textbf{-6.0}} \\
\quad Lexical Only & 38.8 & 33.4 & 36.1 & {\color{red}-3.1} \\
\quad w/o Sentence Smooth & 40.4 & 34.7 & 37.6 & {\color{orange}-1.6} \\
\rowcolor{gray!15}\multicolumn{5}{l}{\textbf{Selection Policy}} \\
\quad Flow Only & 33.3 & 28.8 & 31.1 & {\color{red!80!black}\textbf{-8.1}} \\
\quad Anchor Only & 15.0 & 20.0 & 17.5 & {\color{red!80!black}\textbf{-21.7}} \\
\quad Flash Only & 39.7 & 33.2 & 36.4 & {\color{red}-2.7} \\
\bottomrule
\end{tabular}
}
\caption{Ablation study on LongBench QA (2K budget). $\Delta$ denotes the performance drop relative to \ours.}
\label{tab:component_ablation}
\vspace{-15pt}
\end{table}

\subsection{Model Scalability Analysis}
\label{subsec:scalability}

\begin{figure}
    \centering
    \includegraphics[width=1.0\linewidth]{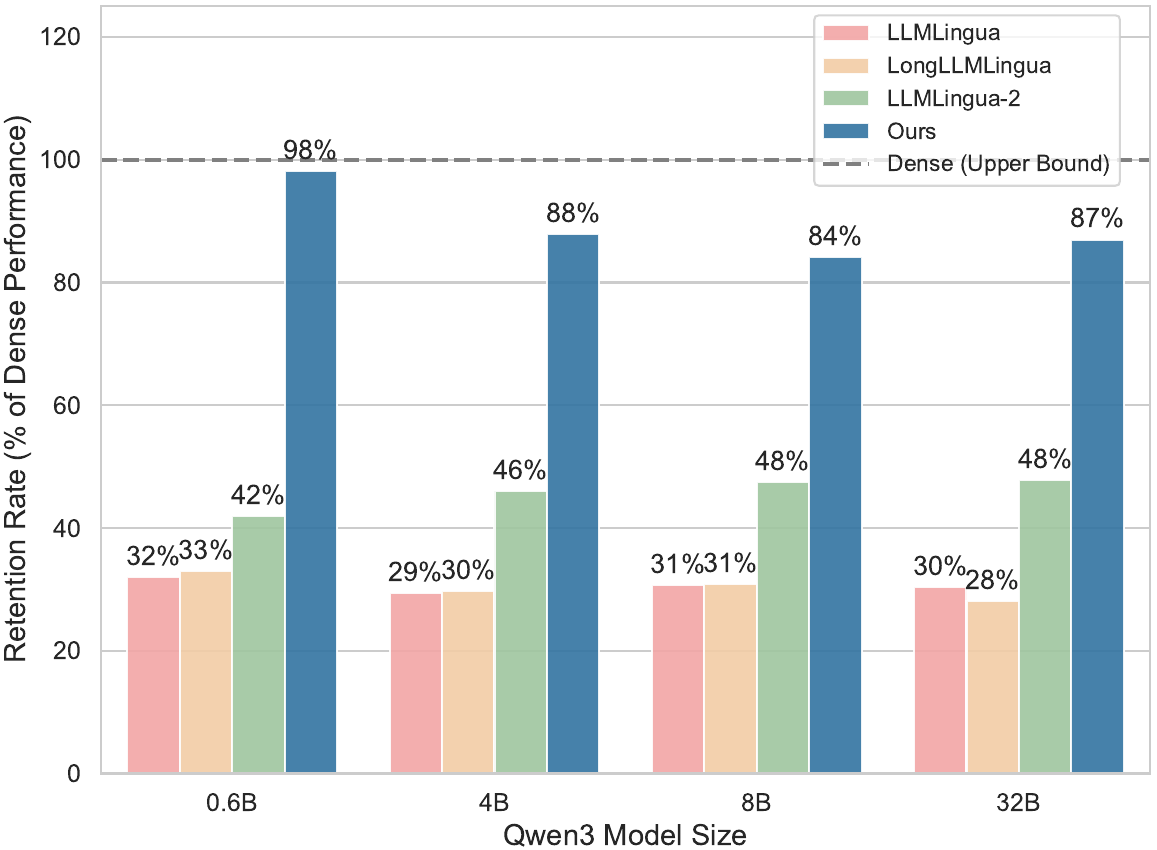}
    \caption{Scalability and robustness analysis on RULER (16k context). Performance is normalized against the Dense upper bound (dashed line represents 100\%).}
    \label{fig:scalability}
    \vspace{-15pt}
\end{figure}

We evaluated the scalability of \ours on the Qwen3 series 0.6B--32B parameters. As shown in Fig.~\ref{fig:scalability}, \ours exhibits high stability across parameter scales compared to learning-based methods, with performance retention consistently between 84\%--98\%. Notably, on the lightweight Qwen3-0.6B, \ours achieves 98\% performance retention, significantly exceeding LLMLingua-2 at 42\% and LongLLMLingua at 33\%. Detailed results are available in Table~\ref{tab:ruler}. This consistent advantage across scales results from the training-free design. By utilizing intrinsic statistical signals instead of fitting specific data distributions, \ours effectively avoids overfitting. This independence from model scale enables \ours to serve as a plug-and-play universal module for seamless integration into various LLM inference pipelines.
More detailed analysis and discussion regarding these results are provided in the Appendix \ref{appendix:scalability}.

\section{Conclusion}
In this paper, we present \ours, a hardware-efficient and training-free prompt compression framework designed to overcome the computational and memory bottlenecks of long-context LLMs. By introducing a hierarchical segment-page mechanism, we transform irregular sequence compression into optimized tensor operations. Our approach leverages unsupervised in-context ITF weighting and a structure-aware hybrid planner to effectively retain both global semantics and fine-grained lexical details, avoiding the semantic collapse common in probability-based pruning. Experimental results confirm that \ours establishes SOTA performance in question answering and fine-grained retrieval, while maintaining highly competitive fidelity in summarization tasks. Crucially, it offers superior scalability, achieving a $26.4\times$ speedup on 128k contexts compared to learning-based baselines. As a plug-and-play module requiring no parameter updates, \ours provides a practical and robust foundation for efficient large-scale long-document understanding.

\section*{Limitations}
Although \ours achieves a favorable balance between efficiency and performance, there are a few limitations.
First, the page-level granularity prioritizes structural integrity and speed, but it is inherently less precise than fine-grained token pruning, which may occasionally result in the retention of minor redundancies within selected segments.
Second, our retrieval mechanism relies on semantic and lexical similarity. It may face challenges in deep multi-hop reasoning scenarios where the supporting evidence shares little direct overlap with the query and requires intermediate deductive steps.
Finally, as a training-free method, \ours depends on pre-set hyperparameters (e.g., weighting factors) that might require manual adjustment to achieve optimal results across drastically different domains, unlike end-to-end learning models that adapt automatically.

\section*{Ethical Considerations}
This work introduces a prompt compression framework, \ours, designed to improve the efficiency of Long Context LLMs. While our method aims to reduce computational costs, there are potential risks associated with information pruning.

The primary risk of any compression method, including \ours, is the potential loss of critical qualifiers or safety constraints (e.g., negations or warnings) present in the original context. In sensitive domains such as legal or medical analysis (as evaluated in our experiments on L-Eval), such omission could lead to factually incorrect or unsafe downstream generations. Although our structure-aware design and sentence smoothing aim to mitigate fragmentation, users should exercise caution and verify outputs when applying this method to high-stakes decision-making tasks.

\ours relies on off-the-shelf embedding models (e.g., Qwen) to calculate page importance. Biases inherent in these pre-trained models, such as the under-representation of certain dialects or cultural contexts, may be propagated. This could result in the systematic filtering of text from minority groups if the embedding model deems such patterns as "low importance" or "redundant." Future work should investigate the fairness of relevance scoring across diverse linguistic demographics.


\bibliography{custom}

\appendix

\begin{table*}[t]
    \centering
    \resizebox{\textwidth}{!}{
    \begin{tabular}{l c c c c c l}
        \toprule
        \textbf{Task Category} & \textbf{\# Subsets} & \textbf{Language} & \textbf{Avg. Length (tokens)} & \textbf{Metric} & \textbf{\# Samples} & \textbf{Representative Datasets} \\
        \midrule
        Single-Doc QA & 4 & 3 En, 1 Zh & 3.6k -- 18.4k & F1 & 750 & NarrativeQA, Qasper, MultiFieldQA \\
        Multi-Doc QA & 4 & 3 En, 1 Zh & 4.9k -- 15.8k & F1 / Rouge-L & 800 & HotpotQA, DuReader, MuSiQue \\
        Summarization & 5 & 4 En, 1 Zh & 2.1k -- 15.4k & Rouge-L & 1,000 & GovReport, VCSUM, SAMSum \\
        Few-Shot Learning & 3 & 2 En, 1 Zh & 5.2k -- 22.3k & Acc. / F1 & 600 & TREC, LSHT, TriviaQA \\
        Synthetic Tasks & 3 & 2 En, 1 Zh & 6.7k -- 11.1k & Accuracy (EM) & 600 & PassageCount, PassageRetrieval \\
        Code Completion & 2 & Multi-Code & 1.2k -- 4.2k & Edit Sim. & 1,000 & LCC, RepoBench-P \\
        \midrule
        \textbf{Total} & \textbf{21} & \textbf{Bi + Code} & \textbf{1.2k -- 22.3k} & \textbf{---} & \textbf{4,750} & \textbf{---} \\
        \bottomrule
    \end{tabular}
    }
    \caption{Statistics of LongBench six major task categories. Here, "En" denotes English datasets, and "Zh" denotes Chinese datasets.}
    \label{tab:longbench_stats}
\end{table*}

\begin{table*}[t]
    \centering
    \resizebox{\textwidth}{!}{
    \begin{tabular}{l c c c c c l}
        \toprule
        \textbf{Task Category} & \textbf{\# Subsets} & \textbf{Language} & \textbf{Avg. Length (words)} & \textbf{Metric} & \textbf{\# Samples} & \textbf{Representative Datasets} \\
        \midrule
        Summarization & 4 & En & 4.9k -- 10.8k & Rouge & 1,378 & GovReport, QMSum, SQuALITY \\
        Question Answering & 4 & En & 1.7k -- 49.4k & F1 / Accuracy & 2,000 & NarrativeQA, Qasper, MuSiQue \\
        Aggregation & 2 & En & 5.5k -- 6.8k & ES / Cidx & 1,000 & SpaceDigest, BookSumSort \\
        \midrule
        \textbf{Total} & \textbf{10} & \textbf{En} & \textbf{1.7k -- 49.4k} & \textbf{---} & \textbf{4,378} & \textbf{---} \\
        \bottomrule
    \end{tabular}
    }
    \caption{Statistics of ZeroSCROLLS task categories. The benchmark consists of 10 datasets across 3 major categories. "ES" denotes Exponential Similarity, and "Cidx" denotes Concordance Index.}
    \label{tab:zero_scrolls_stats}
\end{table*}

\begin{table*}[t]
    \centering
    \resizebox{\textwidth}{!}{
    \begin{tabular}{l c c c c c l}
        \toprule
        \textbf{Task Category} & \textbf{\# Subsets} & \textbf{Language} & \textbf{Length (tokens)} & \textbf{Metric} & \textbf{\# Samples} & \textbf{Representative Datasets} \\
        \midrule
        Retrieval (NIAH) & 8 & En & 4k -- 128k & Accuracy & Variable & S-NIAH, MK-NIAH, KV-Ret \\
        Multi-hop Tracing & 1 & En & 4k -- 128k & Accuracy & Variable & Variable Tracking (VT) \\
        Aggregation & 2 & En & 4k -- 128k & Accuracy & Variable & CWE, FWE \\
        Question Answering & 2 & En & 4k -- 128k & EM / F1 & Variable & SQuAD, HotpotQA \\
        \midrule
        \textbf{Total} & \textbf{13} & \textbf{En} & \textbf{4k -- 128k} & \textbf{---} & \textbf{Variable} & \textbf{---} \\
        \bottomrule
    \end{tabular}
    }
    \caption{Statistics of RULER task categories. RULER contains 13 configurable tasks across 4 categories. The context length is synthetic and typically evaluated from 4k to 128k tokens. "NIAH" denotes Needle In A Haystack tasks.}
    \label{tab:ruler_stats}
\end{table*}

\begin{table*}[t]
    \centering
    \resizebox{\textwidth}{!}{%
    \begin{tabular}{l c c c c c l}
        \toprule
        \textbf{Task Category} & \textbf{\# Subsets} & \textbf{Language} & \textbf{Avg. Length (tokens)} & \textbf{Metric} & \textbf{\# Samples} & \textbf{Representative Datasets} \\
        \midrule
        Summarization & 7 & En & 7.3k -- 20.0k & Rouge & 386 & GovReport, QMSum, SPACE \\
        Open-Ended QA & 6 & En & 3.9k -- 62.3k & Rouge / F1 & 764 & NarrativeQA, NQ, Qasper \\
        Multiple Choice / Exam & 6 & En & 3.9k -- 16.4k & Accuracy & 957 & TOEFL, QuALITY, Coursera \\
        Code Understanding & 1 & En & 31.6k & Accuracy & 90 & CodeU \\
        \midrule
        \textbf{Total} & \textbf{20} & \textbf{En} & \textbf{3.9k -- 62.3k} & \textbf{---} & \textbf{2,197} & \textbf{---} \\
        \bottomrule
    \end{tabular}
    }
    \caption{Statistics of L-Eval Task Categories. The suite comprises 20 datasets categorized by task type. "En" denotes English language tasks.}
    \label{tab:leval_stats}
\end{table*}

\section{Datasets}
\label{appendix:datasets}

\paragraph{LongBench} \cite{bai2024longbench} serves as the first bilingual multitask benchmark designed to evaluate the long context understanding capabilities of Large Language Models (LLMs) on texts ranging from 1k to 22k tokens. As shown in Table~\ref{tab:longbench_stats}, the dataset comprises 21 subdatasets across six task categories involving Chinese, English, and code, totaling 4,750 test samples. All subsets adhere to a standardized input and output format to ensure consistent evaluation across tasks.

\paragraph{ZeroSCROLLS} \cite{shaham2023zeroscrolls} is a zero shot benchmark designed to evaluate natural language understanding capabilities over long texts. It comprises test sets from ten distinct tasks, each requiring reasoning over various types of long content. Unlike traditional benchmarks, ZeroSCROLLS excludes training data to focus on pure zero shot evaluation, requiring models to demonstrate generalization without task specific fine tuning. As shown in Table~\ref{tab:zero_scrolls_stats}, the benchmark covers diverse capabilities such as query based summarization, question answering, multiple choice reasoning, and information aggregation. Each task features natural long texts, with an average document length of approximately 10,000 words.

\paragraph{RULER} \cite{hsieh2024ruler} is a high quality synthetic benchmark designed to evaluate the long context modeling capabilities of language models. Constructed with flexible configurations, it allows for customizable sequence lengths and task complexities to enable comprehensive evaluation across varying context window requirements. The benchmark comprises 13 representative tasks spanning four distinct categories: (i) Retrieval Tasks evaluate the ability to locate specific information and extend beyond the traditional Needle In A Haystack paradigm by covering various needle types and quantities; (ii) Multi hop Tracing Tasks require tracing reasoning chains across segments to test the capacity for maintaining and connecting information throughout long sequences; (iii) Aggregation Tasks assess the capability to collect, summarize, and synthesize information from distributed locations; and (iv) Question Answering Tasks evaluate comprehensive understanding via natural language questions that require synthesizing information from diverse parts of the context. Detailed statistics for RULER are provided in Table~\ref{tab:ruler_stats}.

\paragraph{L-Eval} \cite{an2024eval} L-Eval is a comprehensive benchmark for long context language models designed for the standardized evaluation of processing and reasoning on long inputs. The dataset is constructed through rigorous human annotation and quality inspection to ensure high reliability. It includes 411 documents from diverse domains such as law, finance, academia, and fiction. With an average document length of 7,217 tokens and a total corpus of approximately 60,000 tokens, it is suitable for evaluating long context understanding. The benchmark contains 2,197 manually verified query and answer pairs across 20 tasks. Detailed statistics are presented in Table~\ref{tab:leval_stats}.

\section{Baseline Models}
\label{appendix:baselines}

\begin{table*}[t]
    \centering
    \setlength{\tabcolsep}{2.5pt} 
    \resizebox{\textwidth}{!}{
        \begin{tabular}{l | cccccccccc >{\columncolor{lightgray}}c | cc | cc}
            \toprule
            \multirow{2}{*}{\textbf{Methods}} & \multicolumn{11}{c|}{\textbf{ZeroSCROLLS Datasets}} & \multicolumn{4}{c}{\textbf{Efficiency}} \\
            \cmidrule(lr){2-12} \cmidrule(lr){13-16}
            & GvRp & SSFD & QMsm & SQAL & QALT & Nrtv & Qspr & MuSQ & SpDg & BkSS & \textbf{AVG} & Tokens & $1/\tau$ & Lat. & Spd. \\
            \midrule
            
            \multicolumn{16}{c}{\textit{2,000-token constraint}} \\
            \hline
            
            \multicolumn{16}{l}{\textit{(1) Unsupervised Statistical Methods}} \\
            Selective-Context & 19.0 & 8.4 & 9.7 & 12.4 & 47.0 & 12.5 & 21.6 & 11.5 & 41.2 & 11.0 & 19.4 & 1,865 & 5x & 47.5 & 0.3x \\
            LongLLMLingua     & 20.1 & 12.4 & 14.9 & 16.5 & 65.1 & 27.7 & 30.7 & 23.6 & 68.5 & 47.2 & 32.7 & 1,753 & 6x & 5.2 & 2.3x \\

            \midrule
            \multicolumn{16}{l}{\textit{(2) Supervised and Specialized Learning Methods}} \\
            SBERT             & 10.2 & 7.9 & 13.7 & 13.2 & 60.0 & 8.1 & 10.8 & 1.7 & 37.2 & 42.8 & 20.5 & 1,773 & 6x & 4.1 & 3.0x \\
            OpenAI            & 11.1 & 8.0 & 11.8 & 13.6 & 60.0 & 7.1 & 13.2 & 4.0 & 33.6 & 43.6 & 20.6 & 1,784 & 5x & 9.9 & 1.2x \\
            LLMLingua         & 19.4 & 11.9 & 13.1 & 16.0 & 62.1 & 23.7 & 24.0 & 22.4 & 33.9 & 44.9  & 27.2 & 1,862 & 5x & 4.8 & 2.5x \\
            LLMLingua-2-small & - & - & - & - & - & - & - & - & - & - & \underline{33.3} & 1,862 & 5x & \underline{2.6} & \underline{4.7x} \\
            LLMLingua-2       & - & - & - & - & - & - & - & - & - & - & \textbf{33.4} & 1,898 & 5x & 3.0 & 4.1x \\

            \midrule
            \multicolumn{16}{l}{\textit{(3) Training-free and Zero-Overhead Methods}} \\
            \rowcolor{blue!5} 
            \textbf{\ours (ours)} & 17.2 & 12.5 & 15.2 & 16.5 & 71.4 & 34.0 & 34.4 & 20.2 & 56.8 & 41.8 & 32.0 & 1,878 & 5x & \textbf{2.5} & \textbf{4.9x} \\

            \midrule
            \hline
            
            \multicolumn{16}{c}{\textit{3,000-token constraint}} \\
            \hline

            \multicolumn{16}{l}{\textit{(1) Unsupervised Statistical Methods}} \\
            Selective-Context & 20.8 & 9.1 & 11.7 & 13.4 & 50.0 & 9.8 & 26.1 & 11.0 & 46.0 & 9.5 & 20.7 & 3,460 & 3x & 54.2 & 0.2x \\
            LongLLMLingua     & 22.1 & 12.8 & 15.9 & 17.1 & 67.0 & 27.8 & 31.3 & 23.9 & 65.8 & 46.5 & 33.0 & 3,412 & 3x & 8.2 & 1.5x \\
            
            \midrule
            \multicolumn{16}{l}{\textit{(2) Supervised and Specialized Learning Methods}} \\
            SBERT             & 16.5 & 9.8 & 12.3 & 15.2 & 60.0 & 14.6 & 23.4 & 12.1 & 39.4 & 36.4 & 24.0 & 3,340 & 3x & 5.9 & 2.1x \\
            OpenAI            & 14.3 & 8.3 & 12.0 & 15.3 & 66.7 & 13.3 & 24.3 & 11.7 & 31.2 & 26.4 & 22.4 & 3,362 & 3x & 11.7 & 1.0x \\
            LLMLingua         & 18.7 & 10.0 & 14.9 & 16.8 & 61.9 & 26.9 & 27.2 & 23.4 & 62.9 & 44.5 & 30.7 & 3,366 & 3x & 7.4 & 1.7x \\
            LLMLingua-2-small & - & - & - & - & - & - & - & - & - & - & \underline{33.4} & 3,089 & 3x & \underline{3.0} & \underline{4.1x} \\
            LLMLingua-2       & - & - & - & - & - & - & - & - & - & - & \textbf{33.5} & 3,206 & 3x & 3.5 & 3.5x \\
            
            \midrule
            \multicolumn{16}{l}{\textit{(3) Training-free and Zero-Overhead Methods}} \\
            \rowcolor{blue!5}
            \textbf{\ours (ours)} & 19.4 & 12.5 & 15.9 & 17.5 & 70.6 & 34.0 & 34.5 & 21.8 & 56.0 & 41.2 & 32.4 & 3,319 & 3x & \textbf{2.9} & \textbf{4.2x} \\

            \midrule
            \hline
            \multicolumn{16}{l}{\textit{Reference Baselines}} \\
            Original Prompt   & 21.8 & 12.1 & 17.9 & 17.4 & 66.7 & 25.3 & 29.8 & 20.0 & 69.7 & 44.1 & 32.5 & 9,788 & - & 12.2 & - \\
            Zero-Shot         & 9.4 & 3.0 & 8.6 & 11.4 & 42.9 & 10.6 & 12.4 & 5.5 & 4.2 & 0.0 & 12.8 & 32 & 306x & 1.0 & 12.2x \\
            \bottomrule
        \end{tabular}
    }
    \caption{Detailed evaluation on ZeroSCROLLS datasets. We report the performance on individual sub-tasks: GovReport (GvRp), SummScreenFD (SSFD), QMSum (QMsm), SQuALITY (SQAL), Quality (QALT), NarrativeQA (Nrtv), Qasper (Qspr), MuSiQue (MuSQ), SpaceDigest (SpDg), and BookSumSort (BkSS). "Lat." and "Spd." denote Latency and Speedup, and "-" indicates that the results were not available from the original paper.}
    \label{tab:zeroscrolls_detailed}
\end{table*}

\begin{table*}[h]
    \centering
    \setlength{\tabcolsep}{3pt}
    \resizebox{\linewidth}{!}{
        \begin{tabular}{l | cccccc >{\columncolor{lightgray}}c | cc}
            \toprule
            \multirow{2}{*}{\textbf{Methods}} 
            & \multicolumn{7}{c|}{\textbf{Qwen3-8B}} 
            & \multirow{2}{*}{\textbf{Tokens}$\downarrow$} 
            & \multirow{2}{*}{\bm{$1/\tau$}$\uparrow$} \\
            \cmidrule(lr){2-8}
             & Coursera QA & QuALITY & SFictionQA & TPO & LongFQA & Legal Contract QA & \textbf{AVG} 
             & & \\
            \midrule
            
            LLMLingua 
            & 24.9 & 48.0 & 60.9 & 66.6 & 13.6 & 11.2 & 37.5 
            & 2,122 & 4$\times$ \\

            LongLLMLingua 
            & \textbf{30.2} & 49.0 & 69.5 & 72.9 & 15.2 & 10.9 & 41.3 
            & 2,196 & 4$\times$ \\
            
            LLMLingua-2 
            & 25.3 & \underline{53.0} & \textbf{74.2} & \underline{78.4} & \underline{16.9} & \underline{12.8} & \underline{43.5} 
            & 2,110 & 4$\times$ \\
            
            \midrule
            \rowcolor{blue!5} 
            \textbf{\ours (ours)} 
            & \underline{28.3} & \textbf{57.4} & \underline{73.4} & \textbf{80.3} & \textbf{17.7} & \textbf{13.4} & \textbf{45.1} 
            & 2,180 & 4$\times$ \\
            
            \bottomrule
        \end{tabular}
    }
    \caption{Out-of-domain evaluation on general long-context scenarios (L-Eval) using Qwen3-8B. All methods operate under a 2,000-token input budget constraint. Highest scores are \textbf{bolded}, and second highest are \underline{underlined}.}
    \label{tab:leval-qwen}
\end{table*}

\begin{figure*}[t]
  \centering
  \includegraphics[width=0.8\linewidth]{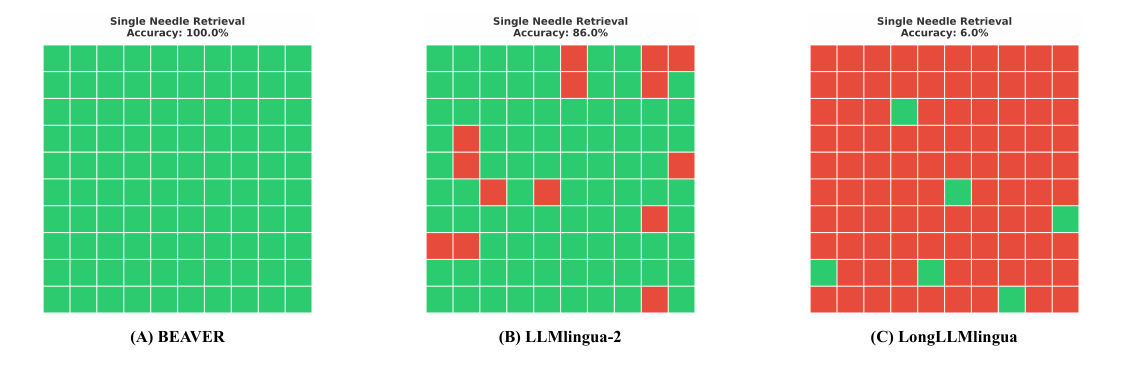}
  \caption{Visualization of Single-Needle Retrieval Accuracy.}
  \label{fig:single-needle}
\end{figure*}

\begin{figure*}[t]
  \centering
  \includegraphics[width=0.8\linewidth]{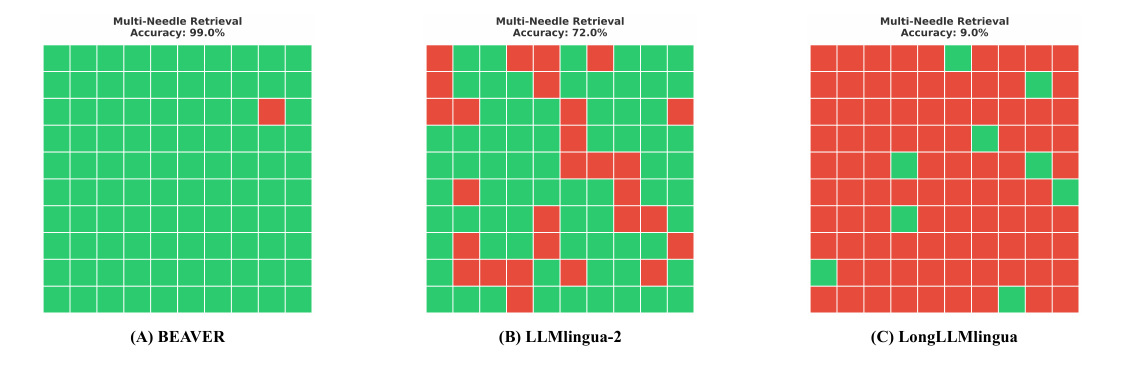}
  \caption{Visualization of Multi-Needle Retrieval Accuracy.}
  \label{fig:multi-needle}
\end{figure*}

\begin{table*}[t]
    \centering
    \renewcommand{\arraystretch}{1.15}
    \setlength{\tabcolsep}{4.5pt}

    \resizebox{\textwidth}{!}{
        \begin{tabular}{l | ccc | cccccc | c | cc | c}
            \toprule
            \multirow{2}{*}{\textbf{Method}} & 
            \multicolumn{3}{c|}{\textbf{Single}} & 
            \multicolumn{6}{c|}{\textbf{Multi}} & 
            \textbf{FWE} & 
            \multicolumn{2}{c|}{\textbf{QA}} & 
            \multirow{2}{*}{\textbf{AVG}} \\
            
            \cmidrule(lr){2-4} \cmidrule(lr){5-10} \cmidrule(lr){11-11} \cmidrule(lr){12-13}
            
             & S-1 & S-2 & S-3 & Key-1 & Key-2 & Key-3 & Val & Qry & VT & Freq & QA-1 & QA-2 & \\
            \midrule
            
            \multicolumn{14}{c}{\textbf{\textit{Qwen3-0.6b}}} \\
            \midrule
            Dense     & \textbf{100.0} & \textbf{100.0} & \textbf{100.0} & \underline{95.0} & \underline{84.0} & \textbf{45.0} & \underline{82.8} & \underline{94.0} & \textbf{86.4} & 90.7 & \textbf{34.0} & \underline{28.0} & \textbf{78.3} \\
            LLMLingua  & \textbf{100.0} & 5.0 & 4.0 & 7.0 & 5.0 & 10.0 & 6.0 & 4.8 & 45.2 & \underline{91.0} & 3.0 & 20.0 & 25.1 \\
            LongLLMLingua  & \textbf{100.0} & 6.0 & 4.0 & 10.0 & 1.0 & 3.0 & 7.5 & 7.0 & 56.6 & \textbf{92.0} & 3.0 & 21.0 & 25.9 \\
            LLMLingua-2  & \underline{27.0} & \underline{69.0} & \underline{24.0} & 57.0 & 7.0 & 2.0 & 44.0 & 45.8 & 0.6 & \textbf{92.0} & 10.0 & 17.0 & 32.9 \\
            \rowcolor{blue!5} 
            \textbf{\ours} & \textbf{100.0} & \textbf{100.0} & \textbf{100.0} & \textbf{97.0} & \textbf{88.0} & \underline{40.0} & \textbf{97.5} & \textbf{97.3} & \underline{58.6} & 89.7 & \underline{25.0} & \textbf{29.0} & \underline{76.8} \\
            \midrule
            
            \multicolumn{14}{c}{\textbf{\textit{Qwen3-4b}}} \\
            \midrule
            Dense     & \textbf{100.0} & \textbf{100.0} & \textbf{100.0} & \textbf{100.0} & \textbf{99.0} & \textbf{100.0} & \textbf{100.0} & \textbf{100.0} & \textbf{99.6} & \textbf{95.7} & \textbf{74.0} & \textbf{55.0} & \textbf{93.6} \\
            LLMLingua  & \textbf{100.0} & 5.0 & 4.0 & 7.0 & 6.0 & 12.0 & 6.0 & 4.8 & 54.2 & \textbf{95.7} & 12.0 & 23.0 & 27.5 \\
            LongLLMLingua  & \textbf{100.0} & 6.0 & 4.0 & 9.0 & 2.0 & 3.0 & 7.8 & 7.0 & 67.8 & \underline{94.3} & 11.0 & 23.0 & 27.9 \\
            LLMLingua-2  & \underline{27.0} & \underline{77.0} & \underline{27.0} & \underline{77.0} & 11.0 & 2.0 & 59.8 & 73.3 & 0.8 & 94.0 & 39.0 & 29.0 & 43.1 \\
            \rowcolor{blue!5} 
            \textbf{\ours} & \textbf{100.0} & \textbf{100.0} & \textbf{100.0} & \textbf{100.0} & \underline{88.0} & \underline{42.0} & \underline{99.8} & \underline{99.8} & \underline{71.8} & 89.7 & \underline{43.0} & \underline{54.0} & \underline{82.3} \\
            \midrule

            \multicolumn{14}{c}{\textbf{\textit{Qwen3-8b}}} \\
            \midrule
            Dense     & \textbf{100.0} & \textbf{100.0} & \textbf{100.0} & \textbf{100.0} & \textbf{100.0} & \textbf{98.0} & \underline{99.0} & \textbf{100.0} & \textbf{100.0} & 93.3 & \textbf{74.0} & \textbf{56.0} & \textbf{93.4} \\
            LLMLingua  & \textbf{100.0} & 5.0 & 4.0 & 7.0 & 6.0 & 12.0 & 6.0 & 4.8 & 54.6 & \textbf{97.0} & 20.0 & 28.0 & 28.7 \\
            LongLLMLingua  & \textbf{100.0} & 6.0 & 4.0 & 10.0 & 1.0 & 3.0 & 8.0 & 7.3 & 62.6 & \underline{96.3} & 20.0 & 29.0 & 28.9 \\
            LLMLingua-2  & \underline{27.0} & \underline{84.0} & \underline{27.0} & \underline{74.0} & 11.0 & 2.0 & 68.5 & 73.5 & 1.8 & 93.3 & 40.0 & 31.0 & 44.4 \\
            \rowcolor{blue!5} 
            \textbf{\ours} & \textbf{100.0} & \textbf{100.0} & \textbf{100.0} & \textbf{100.0} & \underline{84.0} & \underline{32.0} & \textbf{100.0} & \underline{98.8} & \underline{62.8} & 79.0 & \underline{41.0} & \underline{46.0} & \underline{78.6} \\
            \midrule

            \multicolumn{14}{c}{\textbf{\textit{Qwen3-32b}}} \\
            \midrule
            Dense     & \textbf{100.0} & \textbf{100.0} & \textbf{100.0} & \textbf{100.0} & \textbf{100.0} & \textbf{100.0} & \underline{99.5} & \textbf{100.0} & \textbf{100.0} & \textbf{97.7} & \textbf{86.0} & \textbf{60.0} & \textbf{95.3} \\
            LLMLingua  & \textbf{100.0} & 5.0 & 4.0 & 7.0 & 6.0 & 12.0 & 6.0 & 4.8 & 60.0 & \underline{97.3} & 18.0 & 28.0 & 29.0 \\
            LongLLMLingua  & \textbf{100.0} & 6.0 & 4.0 & 10.0 & 1.0 & 3.0 & 8.0 & 7.3 & 36.0 & 96.3 & 22.0 & 28.0 & 26.8 \\
            LLMLingua-2  & \underline{27.0} & \underline{82.0} & \underline{27.0} & \underline{74.0} & 9.0 & 2.0 & 68.0 & 75.8 & 3.2 & 94.7 & 49.0 & 35.0 & 45.6 \\
            \rowcolor{blue!5} 
            \textbf{\ours} & \textbf{100.0} & \textbf{100.0} & \textbf{100.0} & \textbf{100.0} & \underline{88.0} & \underline{41.0} & \textbf{100.0} & \underline{99.8} & \underline{70.0} & 90.0 & \underline{51.0} & \underline{55.0} & \underline{82.9} \\
            
            \bottomrule
        \end{tabular}
    }
    \caption{Main Results on RULER benchmark (16k context).
    Highest scores are \textbf{bolded}, and second highest are \underline{underlined}.
    }
    \label{tab:ruler}
\end{table*}

\subsection{Unsupervised Statistical Methods}

\paragraph{Selective-Context}\cite{li2023selective} proposes a context compression technique designed to enhance LLM inference efficiency by identifying and pruning redundant input information. The method employs a base causal language model to evaluate the self-information of lexical units (such as sentences or tokens), selectively retaining content with higher information values while discarding predictable redundancy.

\paragraph{LongLLMLingua}\cite{jiang2024longllmlingua} addresses the challenges of high computational cost, noise, and position bias inherent in long-context scenarios. Building upon LLMLingua, it introduces a question-aware coarse-to-fine compression method to enhance the density of key information relative to the user's query. Furthermore, the method incorporates a document reordering strategy to mitigate the "lost-in-the-middle" phenomenon and employs a post-compression subsequence recovery mechanism to preserve the integrity of critical details.

\subsection{Supervised and Specialized Learning Methods}

\paragraph{LLMLingua}\cite{jiang2023llmlingua} presents a coarse-to-fine prompt compression framework designed to accelerate inference by pruning prompts while preserving semantic integrity. The method employs a budget controller to dynamically allocate compression ratios across different prompt components and utilizes a token-level iterative algorithm to capture conditional dependencies between tokens. Additionally, it incorporates instruction tuning to align the distribution of the smaller compression model with the target LLM.

\paragraph{Sentence-BERT}\cite{reimers2019sentence} modifies the pretrained BERT architecture by employing siamese and triplet network structures to derive semantically meaningful sentence embeddings. This framework allows sentences to be compared using cosine similarity, thereby significantly reducing the computational overhead associated with pair-wise cross-encoding while maintaining high performance on semantic search and clustering tasks.

\paragraph{LLMLingua-2}\cite{pan2024llmlingua2} advances task-agnostic prompt compression by framing the problem as token classification rather than causal entropy-based pruning. The authors distill knowledge from GPT-4 to create a high-quality extractive compression dataset, which is then used to train a bidirectional Transformer encoder (e.g., XLM-RoBERTa, mBERT) to classify tokens as "preserve" or "discard". This approach enables the model to leverage full bidirectional context for better information retention, ensuring faithfulness to the original text while achieving significantly lower latency compared to autoregressive or causal-based methods.

\section{Detailed Analysis on ZeroSCROLLS}
\label{appendix:details_zero}

Table~\ref{tab:zeroscrolls_detailed} reports the detailed results on the ZeroSCROLLS benchmark. Overall, our proposed \ours framework demonstrates strong robustness across all ten subtasks, achieving average scores of 32.0 and 32.4 under the compression budgets of 2,000 and 3,000 tokens, respectively. This performance substantially surpasses unsupervised statistical baselines such as Selective-Context and achieves results comparable to LongLLMLingua. Notably, although \ours is entirely training-free, it matches or even exceeds state-of-the-art supervised methods such as LLMLingua-2 across metrics. For example, it maintains highly competitive scores on GovReport and SQuALITY while avoiding the costly overhead of model fine-tuning. On detail-sensitive tasks such as Qasper and QMSum, \ours still attains high accuracy, largely because it can identify key cues such as entities and numbers that are often overlooked in purely semantic spaces. Moreover, on tasks requiring long-range dependency understanding, such as NarrativeQA, the model achieves a high score of 34.0. This result supports the necessity of the structural priors introduced in our QueryPlanner, where retaining anchor pages and local streaming context effectively preserves discourse coherence and causal structure.

\section{Needle-in-a-Haystack Visualization}
\label{appendix:needle}

To provide a comprehensive evaluation of context modeling capabilities, we visualize the retrieval robustness of \ours compared to baselines using the Needle-in-a-Haystack test on the RULER benchmark. We conduct experiments under a high-compression setting. Specifically, the original context length is fixed at 16k tokens, while the compressed token constraint is strictly limited to 3,000 tokens (approximately $5.3\times$ compression). We evaluate both Single-Needle and Multi-Needle retrieval tasks across 100 distinct samples. All inference is performed using \texttt{gpt-3.5-turbo} as the backend model to ensure consistent evaluation.

Figures~\ref{fig:single-needle} and \ref{fig:multi-needle} present the retrieval heatmaps, where green blocks indicate successful retrieval and red blocks denote failure. The results demonstrate the superior robustness of \ours, which achieves near-perfect performance by maintaining a "full green" heatmap with 100.0\% and 99.0\% accuracy in Single-Needle and Multi-Needle retrieval, respectively. This indicates that \ours effectively preserves semantic integrity even under aggressive compression. In contrast, LLMlingua-2 exhibits degradation in complex scenarios; while retaining acceptable performance in the single-needle task (86.0\%), its accuracy declines significantly to 72.0\% in the multi-needle setting, reflecting an inability to consistently retrieve distributed information. Furthermore, LongLLMlingua displays severe information loss characterized by predominantly red heatmaps, achieving only 6.0\% and 9.0\% accuracy for single and multi-needle tasks, which suggests it struggles to prioritize critical query-relevant information within the 3,000-token budget.

\section{Additional Evaluation on L-Eval Benchmark}
\label{appendix:leval_results}

To further verify the generalization capability of \ours on open-weights models beyond the ZeroSCROLLS and RULER benchmarks, we conducted an out-of-domain evaluation on the L-Eval suite \cite{an2024eval} using \texttt{Qwen3-8B}\footnote{\url{https://huggingface.co/Qwen/Qwen3-8B}}. L-Eval provides a diverse set of long-context tasks including academic QA, legal analysis, and fiction reading, which serve as a robust testbed for real-world applicability.

The results are presented in Table~\ref{tab:leval-qwen}. Consistent with our observations on GPT-series models, \ours demonstrates superior performance on Qwen3-8B, achieving the highest average score of 45.1. Our method outperforms the strongest baseline, LLMLingua-2, by approximately 1.6 points on average. Notably, in complex reasoning tasks such as \textit{TPO} (Topology) and \textit{Legal Contract QA}, \ours leads by significant margins (+1.9 and +0.6 respectively), indicating that our structure-aware compression better preserves logical dependencies required for specialized domains. All methods operate under a similar compression ratio (approx. $4\times$, reducing context to 2,000 tokens). \ours achieves this performance gain without the need for external training (unlike LLMLingua-2) or unstable perplexity calculations (unlike LongLLMLingua), highlighting its effectiveness as a lightweight, plug-and-play solution for open-source LLMs.

\section{Qualitative Comparison with Baseline Methods}
\label{appendix:qualitative_comparison}

To provide an intuitive assessment of our method under strict inference constraints, we conducted qualitative analyses on four representative tasks from the L-Eval benchmark \cite{an2024eval}, including code generation (CodeU), government report summarization (GovReport \cite{huang2021efficient}), financial question answering (LongFQA), and few shot mathematical reasoning (GSM100). We impose strict length budgets ranging from 500 to 1500 tokens to evaluate each method's ability to preserve critical information under extreme compression. We compared \ours with unsupervised statistical methods (for example, LongLLMLingua \cite{jiang2024longllmlingua}) and supervised methods (for example, LLMLingua 2 \cite{pan2024llmlingua2}), and focus on failure modes related to semantic noise, program logic, chain of thought structure, and code syntax.

\begin{figure*}[t]
  \centering
  \includegraphics[width=\linewidth]{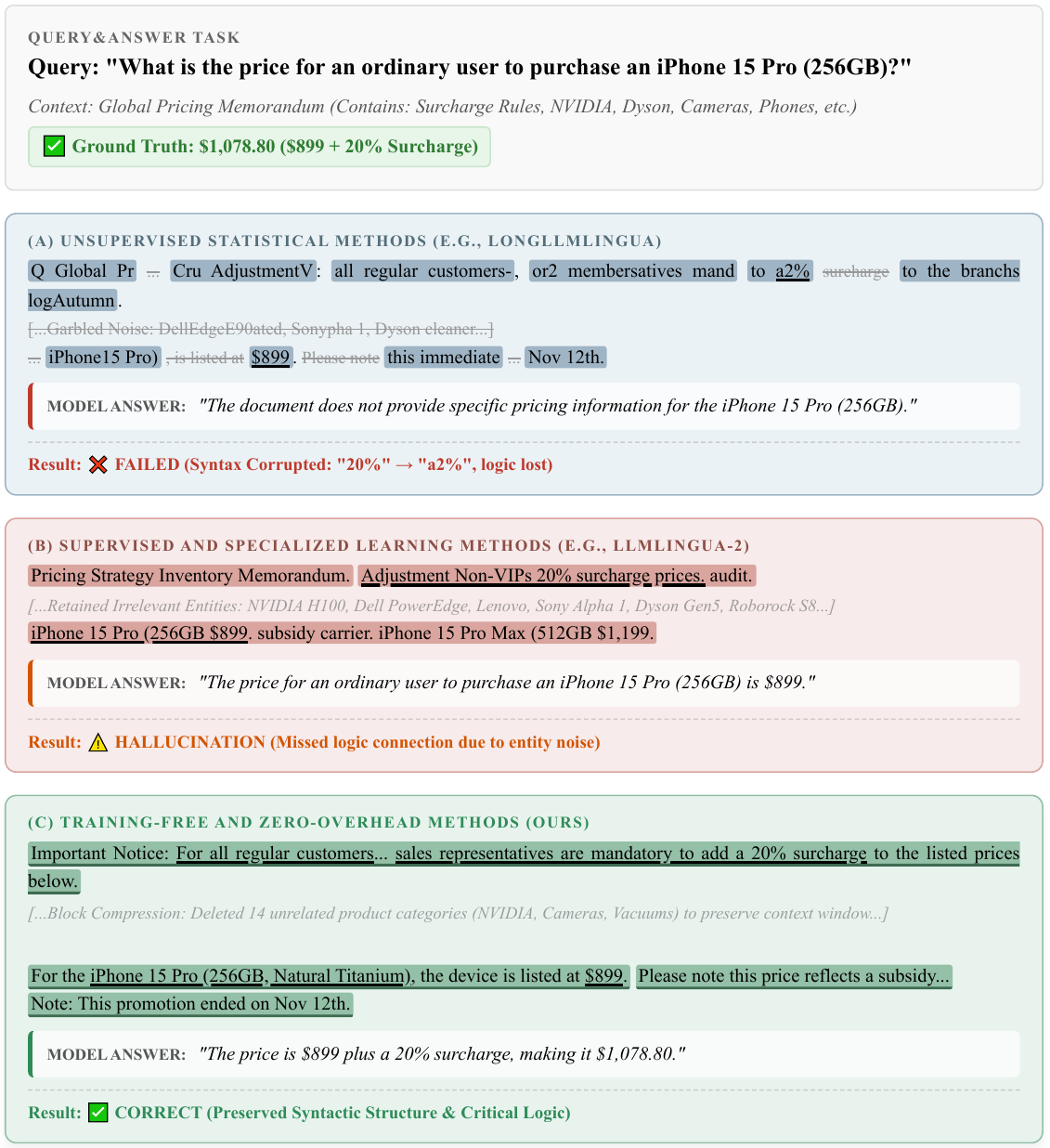}
  \caption{Qualitative comparison on Financial QA under a 500 token budget.}
  \label{fig:qual_financial_qa}
\end{figure*}

First, in LongFQA with substantial semantic noise under a 500 token budget, the model must extract the pricing logic of iPhone 15 Pro from context mixed with irrelevant inventory information about consumer electronics such as NVIDIA GPUs and Dyson vacuum cleaners (Figure~\ref{fig:qual_financial_qa}). Results show that the statistical baseline (Baseline A) over compresses and corrupts critical surface forms, for example deleting the key surcharge term ``20\%'' into ``a2\%'', which breaks the numerical logic. The supervised baseline (Baseline B) produces fluent text but fails to preserve the causal relation between the base price and the surcharge rule, leading to hallucination and an incorrect output of \$899. In contrast, our method uses a hybrid semantic and lexical scoring mechanism to identify and retain the price table and the surcharge memo scattered across the document, while filtering the intervening noise blocks. This enables the downstream model to correctly infer the final price of \$1{,}078.80, demonstrating strong robustness to distraction.

\begin{figure*}[t]
  \centering
  \includegraphics[width=\linewidth]{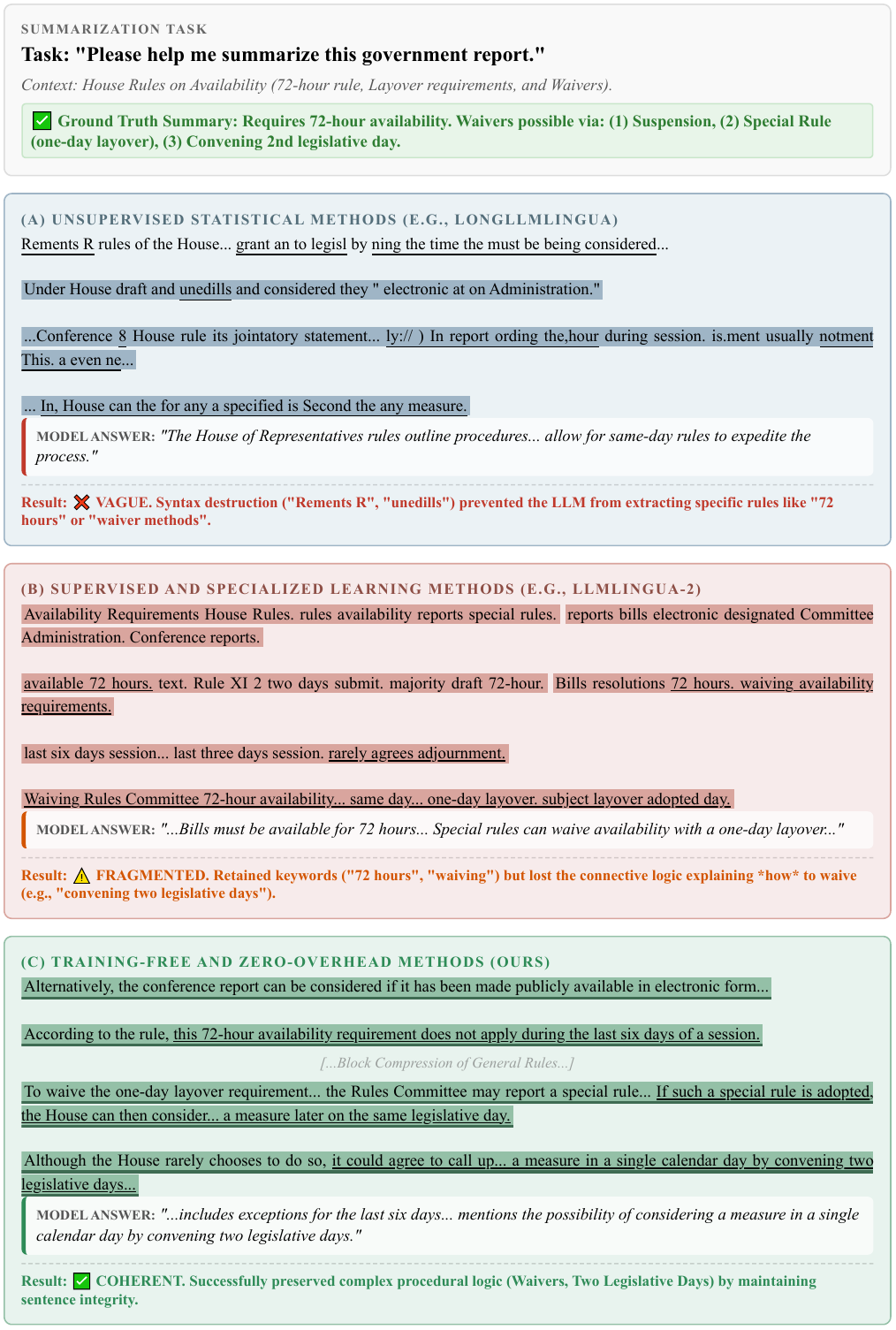}
  \caption{Qualitative comparison on GovReport summarization under a 500 token budget.}
  \label{fig:qual_govreport}
\end{figure*}

Second, for the GovReport summarization task under a 500 token budget, which tests long document coherence and logical completeness, the goal is to extract the procedural logic of the ``72 hour rule'' and its exemption conditions (Figure~\ref{fig:qual_govreport}). Baselines exhibit severe semantic fragmentation. The statistical method damages syntactic structure, for example generating corrupted fragments such as ``Rements R'', which prevents rule extraction. The supervised method retains keywords such as ``one day layover'' but loses the connecting logic that triggers the exemption, namely ``convening two legislative days'', making the summary factually incomplete. By contrast, our method leverages a Segmenter to respect natural language boundaries, preserving complete sentence structure for both the rule and its exceptions, and yields a coherent and accurate summary.

\begin{figure*}[t]
  \centering
  \includegraphics[width=\linewidth]{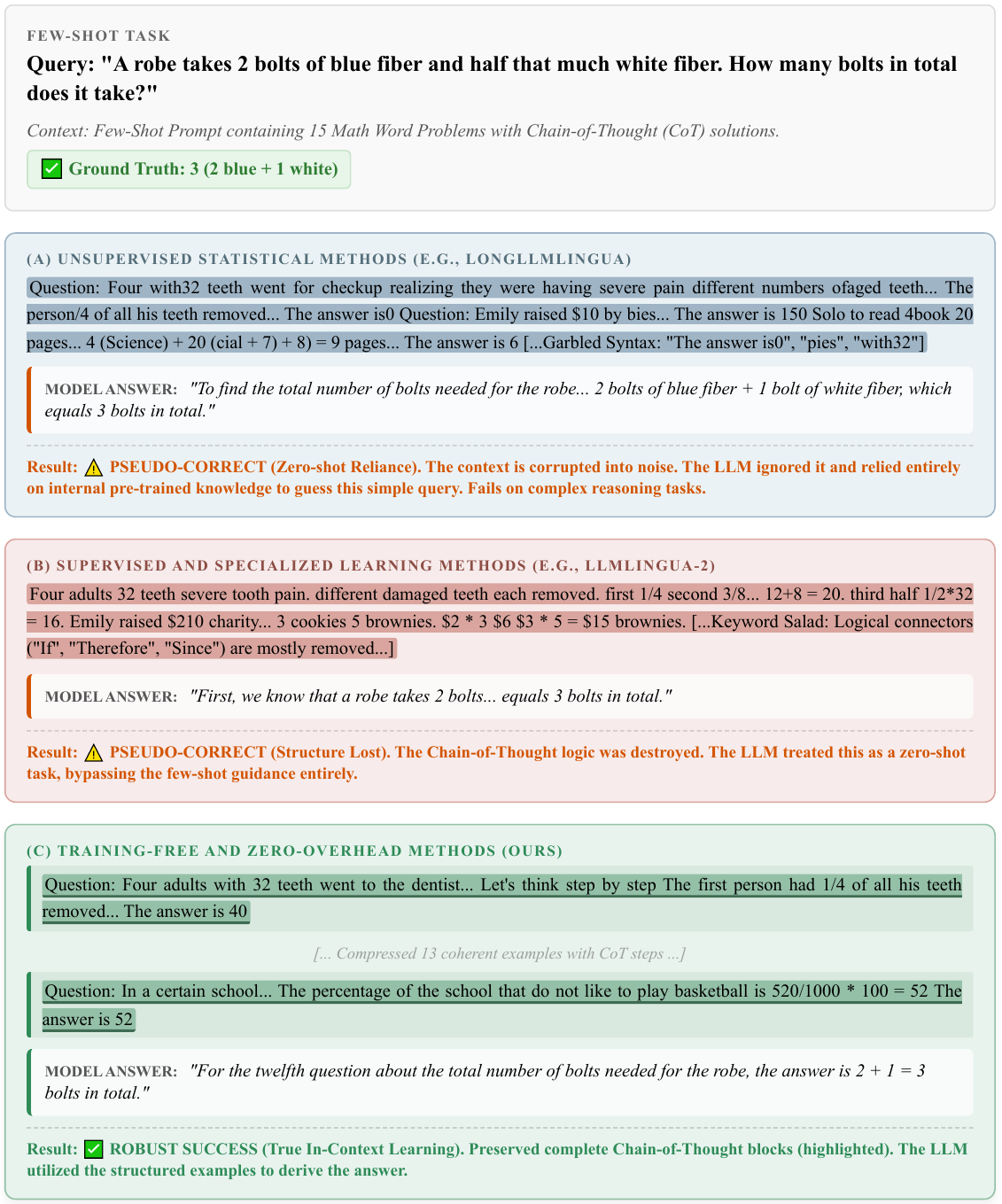}
  \caption{Qualitative comparison on GSM100 under a 1500 token budget.}
  \label{fig:qual_gsm100}
\end{figure*}

Third, on the GSM100 few shot reasoning task with a 1500 token budget, we evaluate whether compression preserves chain of thought structure (Figure~\ref{fig:qual_gsm100}). Baselines often treat few shot demonstrations as unstructured text and destroy the boundaries between question, reasoning, and answer, causing the large model to miss the in context examples and degrade to zero shot guessing, which fails on nontrivial arithmetic. Our method treats each example as a coherent page unit and preserves the chain of thought steps of highly relevant demonstrations, such as ``Let's think step by step...'', enabling genuine in context learning and guiding the model to the correct result $2+1=3$.

\begin{figure*}[t]
  \centering
  \includegraphics[width=\linewidth]{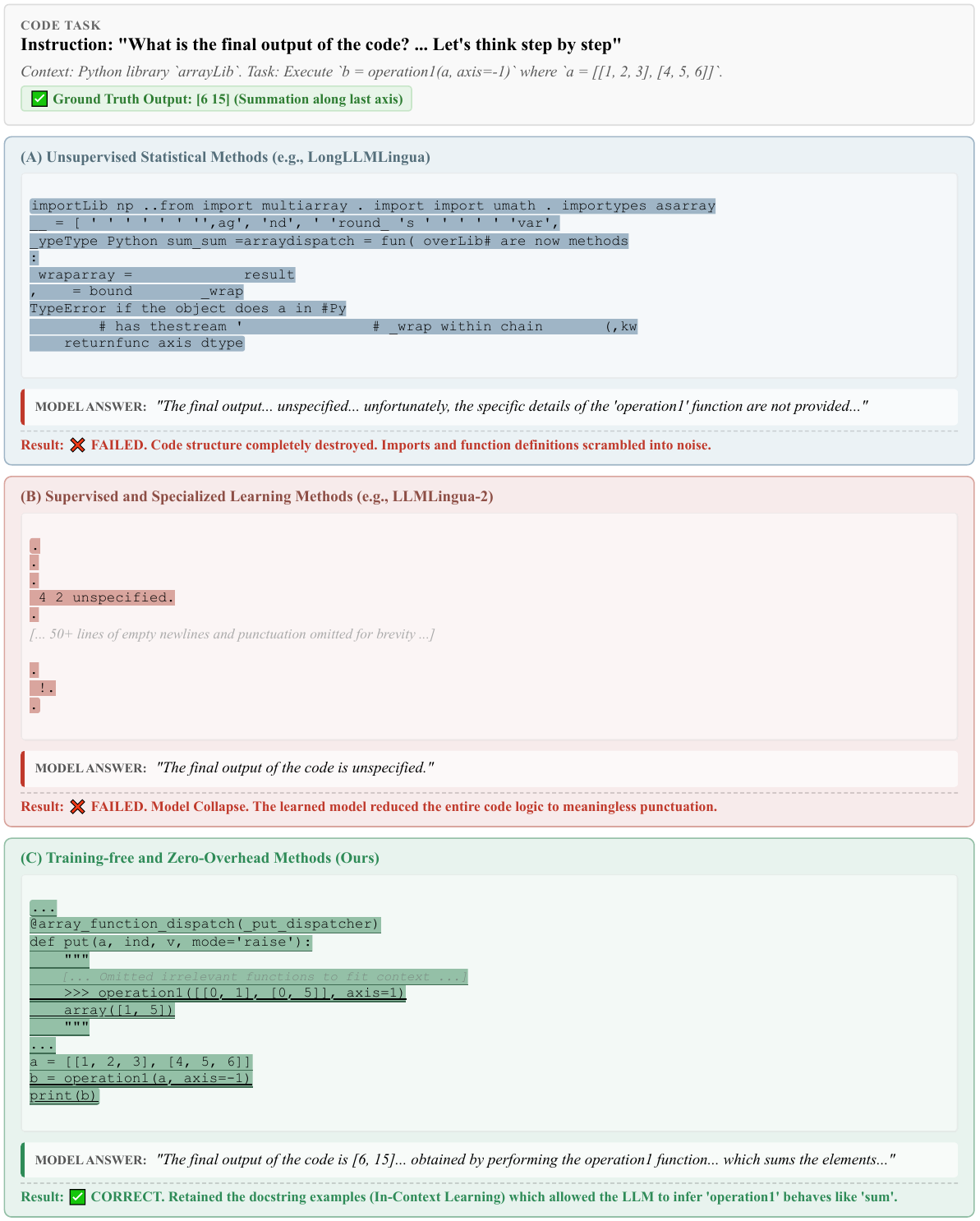}
  \caption{Qualitative comparison on CodeU under a 1000 token budget.}
  \label{fig:qual_codeu}
\end{figure*}

Finally, in the CodeU code task under a 1000 token budget, where syntactic correctness is critical, baselines struggle more severely (Figure~\ref{fig:qual_codeu}). The statistical method indiscriminately removes tokens deemed redundant, including structural delimiters, yielding unparsable code such as ``importLib np..''. The supervised method can collapse, compressing complex logic into meaningless punctuation. Given the high information density of code, our method prioritizes rare tokens and structural separators through the lexical pathway in PageEncoder, preserving function definitions and the demonstration examples in docstrings. This allows the model to pass syntax checks and to learn from examples to correctly predict execution outcomes.

\section{Additional Analysis on Scalability and Robustness}
\label{appendix:scalability}

In this section, we provide a detailed analysis of the scalability results presented in Table~\ref{tab:ruler}, examining the performance of \ours across the Qwen3 series (0.6B to 32B) in comparison to the two categories of baselines defined in the related work: Unsupervised Statistical Methods and Supervised Specialized Learning Methods.

\subsection{Analysis vs. Unsupervised Statistical Methods}
Unsupervised methods like LongLLMLingua rely on causal entropy or perplexity metrics to identify key information. While these metrics are effective for larger models, our experiments reveal their fragility on lightweight architectures. As shown in Table~\ref{tab:ruler}, LongLLMLingua achieves an average score of only 25.9 on Qwen3-0.6B. This performance drop indicates that probability distributions in smaller models are significantly noisier and less reliable as proxies for semantic importance compared to their larger counterparts. Furthermore, the "question-aware" mechanism in LongLLMLingua depends on the model's zero-shot understanding of the query, which is inherently limited in 0.6B-scale models. In contrast, \ours utilizes intrinsic attention statistics, achieving a score of \textbf{76.8} on the same 0.6B model. This substantial margin demonstrates that attention-based signals are far more robust and invariant to model scale than the perplexity-based signals used by other unsupervised baselines.

\subsection{Analysis vs. Supervised and Specialized Learning Methods}
For supervised methods such as LLMLingua and LLMLingua-2, the primary bottleneck on smaller models stems from the distribution mismatch between the compression module and the target inference model. LLMLingua-2, for instance, employs a BERT-based encoder trained on data distilled from GPT-4. While this effectively captures information relevant to GPT-4, these patterns do not transfer well to the Qwen3-0.6B model (score: 32.9). The tokens considered "informative" by a GPT-4-aligned compressor are often too complex or distinct from the reasoning paths required by a lightweight model. Similarly, LLMLingua's instruction tuning strategy fails to align effectively when the target model lacks sufficient capacity, resulting in the lowest score of 25.1. \ours circumvents these compatibility issues by design. By deriving compression decisions directly from the specific model instance being used, \ours ensures that the retained information is inherently aligned with that specific model's processing mechanisms. This "self-adaptive" characteristic allows \ours to maintain 98\% performance retention on the 0.6B model without the need for external training or domain adaptation.

\section{LLM Usage}
\label{appendix:llm_usage}

We acknowledge the use of LLMs to assist with the writing process of this manuscript. Specifically, these tools were employed solely for the purpose of refining textual expression, correcting grammatical errors, and improving the flow of the narrative.

We explicitly state that:
\begin{itemize}
    \item The conceptualization, methodology design, and analysis of experimental results are entirely the original work of the authors.
    \item No part of the experimental code, data processing pipelines, or implementation details was generated by AI assistants.
\end{itemize}

\end{document}